%% file: main.tex
\documentclass[10pt,twocolumn,letterpaper]{article}
\usepackage[pagenumbers]{cvpr}
\input{preamble}
\title{Bridging Update Coordination and Flatness in Federated Learning}
\input{authors}
\hypersetup{
  pdftitle={Bridging Update Coordination and Flatness in Federated Learning},
  pdfauthor={Chongwei He, Shichen Huang, Jiayun Li, Deyuan Lin, Wangze Jiang, Shengjia Li, Zhi-Ri Tang, Xiaofan Li}
}
\begin{document}
\maketitle
\input{sections/00_abstract}
\input{sections/01_introduction}
\input{sections/02_related_work}
\input{sections/03_method}
\input{sections/04_experiments}
\input{sections/05_conclusion}

\section*{LLM Usage Statement}
OpenAI Codex assisted with drafting, translation, and editing of the manuscript, including the abstract, introduction, related work, method, and experimental presentation. It also assisted with implementation and analysis scripts, preparation of figures and tables, and \LaTeX\ formatting. The numerical results reported in the paper are based on experiment records.

\par
{\small
\bibliographystyle{ieeenat_fullname}
\bibliography{references}
}

\clearpage
\appendix
\section*{Appendix}
\setcounter{equation}{0}
\setcounter{table}{0}
\setcounter{algorithm}{0}
\renewcommand{\theequation}{S\arabic{equation}}
\renewcommand{\thetable}{S\arabic{table}}
\renewcommand{\thealgorithm}{S\arabic{algorithm}}
\renewcommand{\theHequation}{appendix.\arabic{equation}}
\renewcommand{\theHtable}{appendix.\arabic{table}}
\renewcommand{\theHalgorithm}{appendix.\arabic{algorithm}}
\input{sections/supplementary_content}

\end{document}

%% file: preamble.tex
\usepackage{times}
\usepackage{graphicx}
\usepackage{amsmath,amssymb,amsthm}
\DeclareMathSizes{5.8}{5.8}{5}{5}
\DeclareMathSizes{5.2}{5.2}{5}{5}
\usepackage{booktabs,array,multirow}
\newcommand{\resultspanhead}[1]{%
  \raisebox{\dimexpr(\depth-\height-\aboverulesep-\cmidrulewidth-\belowrulesep)/2-\dp\strutbox\relax}[0pt][0pt]{%
    \shortstack[c]{#1}}}
\usepackage{algorithm}
\usepackage{algpseudocode}
\usepackage{xcolor}
\usepackage{colortbl}
\usepackage{microtype}
\definecolor{cvprblue}{rgb}{0.21,0.49,0.74}
\usepackage[pagebackref,breaklinks,colorlinks,allcolors=cvprblue]{hyperref}
\usepackage[capitalize,noabbrev]{cleveref}

\newcommand{\E}{\mathbb{E}}
\newcommand{\norm}[1]{\left\lVert #1\right\rVert}
\newcommand{\coord}{\mathcal{A}}
\newcommand{\probe}{\mathcal{P}}
\newcommand{\update}{\mathcal{U}}
\newcommand{\hist}{\mathcal{S}}

\newcommand{\shared}{\mathrm{s}}
\newcommand{\loc}{\mathrm{loc}}

\newcommand{\macroacc}{A_{\mathrm{macro}}}
\newcommand{\minacc}{A_{\min}}

\DeclareMathOperator*{\argmin}{arg\,min}

\DeclareMathOperator{\diag}{diag}

\newtheorem{lemma}{Lemma}
\crefname{proposition}{Proposition}{Propositions}
\crefname{lemma}{Lemma}{Lemmas}
\crefname{algorithm}{Algorithm}{Algorithms}
\newcolumntype{L}[1]{>{\raggedright\arraybackslash}p{#1}}

%% file: authors.tex
\author{%
\begin{tabular}{@{}c@{\quad}c@{\quad}c@{\quad}c@{\quad}c@{}}
Chongwei He$^{1}$\thanks{Equal contribution.} & Shichen Huang$^{1*}$ & Jiayun Li$^1$ & Deyuan Lin$^1$ & Wangze Jiang$^1$\\
\multicolumn{5}{@{}c@{}}{Shengjia Li$^2$\quad Xu Wang$^1$\quad Zhi-Ri Tang$^1$\thanks{Corresponding authors: Zhi-Ri Tang and Xiaofan Li.\\\mbox{tangzhiri@jnu.edu.cn}, \mbox{lixiaofan@jnu.edu.cn}.}\quad Xiaofan Li$^{1\dagger}$}
\end{tabular}\\[7pt]
\begin{minipage}[t]{0.50\textwidth}
\centering
$^1$School of Intelligent Science and Engineering /\\
School of Artificial Intelligence Industry\\
Jinan University Zhuhai Campus\\
Zhuhai 519070, China
\end{minipage}%
\hspace{0.02\textwidth}%
\begin{minipage}[t]{0.44\textwidth}
\centering
$^2$School of Intelligent Civil\\
and Ocean Engineering\\
Harbin Institute of Technology, Shenzhen\\
Shenzhen 518055, China
\end{minipage}
}

%% file: sections/00_abstract.tex
\begin{abstract}
Federated learning combines complementary knowledge across clients to alleviate local data limitations without sharing raw samples. However, data heterogeneity can cause conflicting updates and hinder generalization. Update coordination reduces interference among client updates, while flatness optimization reduces sensitivity to parameter perturbations to improve generalization. Existing joint designs largely target specific algorithms, limiting flexible combinations of these approaches. To address this problem, we propose a unified federated learning framework that bridges update coordination and flatness optimization. Clients separately accumulate descent signals and flatness corrections along the same local trajectory. The server coordinates both groups with independent states before fusing their outputs into a global update. Explicit conversion between trajectory signals and model updates enables different coordination strategies to work with compatible local or globally guided flatness mechanisms without changing the inference architecture. Experiments show that the evaluated combinations generally improve average and worst-client accuracy over update coordination alone, demonstrating the effectiveness of the proposed framework in combining these complementary optimization mechanisms.
\end{abstract}

%% file: sections/01_introduction.tex
\section{Introduction}
\label{sec:intro}
\input{sections/01_bridge_figure}

Federated learning enables clients to benefit from distributed data through knowledge sharing without centralizing raw samples~\cite{mcmahan2017fedavg}. However, differences in client data make collaborative training challenging. For example, in cross-domain object recognition, clients may hold artistic images, clip art, product images, and real-world photographs, as represented by the four domains of Office-Home~\cite{venkateswara2017officehome}. Although they recognize the same object categories, differences in visual style and background can lead clients to learn different features~\cite{li2021fedbn}. An update that benefits one client may interfere with others~\cite{hu2022fedmgda}, while overreliance on domain-specific features can limit generalization. Federated training therefore needs both coordinated client updates and improved shared-model generalization.

Update coordination and flatness optimization offer complementary perspectives on this problem. We use \emph{update coordination} to denote operations that adjust client directions or their aggregation weights to organize the shared update~\cite{wang2021fedfv,hu2022fedmgda,shi2025fedawa}. Flatness optimization uses parameter perturbations to seek low-loss regions that are less sensitive to nearby parameter changes, with the aim of improving generalization~\cite{foret2021sam,qu2022fedsam}. These perspectives address how clients contribute to a shared update and how the model responds to parameter changes. Their joint use offers an opportunity to improve collaboration and predictive performance.

Joint methods link flatness to local proximal correction~\cite{sun2023fedspeed} or to global state through shared perturbations~\cite{sun2023fedsmoo} and server momentum~\cite{li2025fedwmsam}. Supporting independent choices of these components introduces two challenges. First, descent and flatness signals must remain distinguishable over multiple local steps, with an explicit conversion to model updates even when clients take different numbers of steps. Second, a coordinator's historical state or adaptive weights must retain a clear meaning when applied to each signal. Mixing both signals into one vector obscures their individual cross-client relationships, while reusing a single state entangles their histories.

To address these challenges, we propose a unified federated learning framework (\cref{fig:bridge}). The framework serves as a bridge between update coordination and flatness optimization, enabling their mechanisms to be combined within a common training procedure. Clients accumulate ordinary gradients and perturbation-induced corrections along one local trajectory. An endpoint reconstruction identity relates these signals to the realized model update, and adapters translate between the representations required by different coordinators. The server coordinates the two groups using independent states, then fuses their outputs for a shared-model update. Thus, the flatness mechanism, coordination operator, and server optimizer have explicit, separate roles.

Compatible coordination and flatness components can be replaced independently without changing the inference architecture. The framework retains branch-specific relationships and histories for controlled composition and analysis. Experiments on benchmarks including Fashion-MNIST show that the evaluated component combinations generally improve performance over their corresponding ordinary-update baselines, demonstrating the value of combining these complementary optimization strategies within a shared framework.

Our contributions are as follows.
\begin{itemize}
    \item We propose a unified federated learning framework that bridges update coordination and flatness optimization. It accumulates both signals along one local trajectory and isolates their coordination states, allowing compatible components to be combined and replaced independently.
    \item We characterize the complementary roles of descent and curvature, establish exact reconstruction of model updates from trajectory signals, and prove that separate and direct coordination are equivalent under the same fixed linear aggregation rule.
    \item Across benchmarks, the evaluated combinations generally improve average and worst-client accuracy. A controlled comparison shows that separate coordination achieves accuracy comparable to direct coordination while retaining independent branch states.
\end{itemize}

%% file: sections/01_bridge_figure.tex
\begin{figure*}[t]
    \centering
    \includegraphics[width=\textwidth]{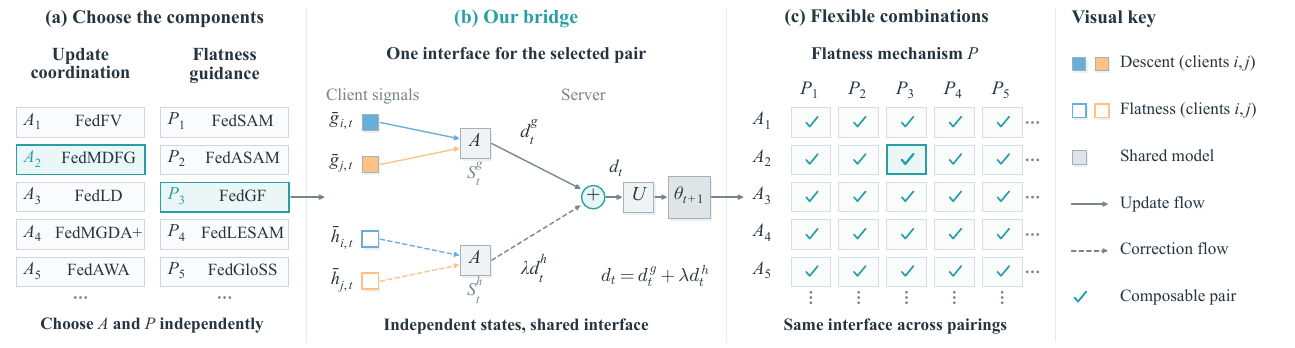}
    \caption{\textbf{The bridging role of the proposed framework.}
    (a) A coordination strategy $A$ and a flatness mechanism $P$ are selected independently through their adapters. FedSAM and FedASAM denote the local SAM-guided and ASAM-guided federated variants.
    (b) The server coordinates descent and flatness signals with independent states, fuses the outputs, and applies the shared-model update through $U$.
    (c) The framework supports different pairings through the same coordination interface, with FedMDFG and FedGF highlighted as an example. Checks indicate adapter-enabled compositions, and ellipses indicate extensibility beyond the listed examples.}
    \label{fig:bridge}
\end{figure*}

%% file: sections/02_related_work.tex
\section{Related Work}
\label{sec:related}
\subsection{Update coordination in federated learning}
\label{sec:related_coordination}
Aggregation-stage coordination addresses conflicts among first-order client directions. FedProx~\cite{li2020fedprox} and SCAFFOLD~\cite{karimireddy2020scaffold} address local drift via proximal regularization and control variates, respectively. At aggregation, FedFV~\cite{wang2021fedfv} projects conflicting updates, while CRAFT~\cite{wang2026craft} projects a reference direction under alignment constraints, also at the layer level. FedLF~\cite{pan2024fedlf} and FedCALM~\cite{zheng2025fedcalm} treat conflicts layer by layer, with FedCALM designed for selective aggregation in personalized models. FedMGDA+~\cite{hu2022fedmgda} and FedMDFG~\cite{pan2023fedmdfg} seek common descent via minimum-norm optimization over normalized directions and fairness-guided optimization, respectively. FedAWA~\cite{shi2025fedawa}, FedPW~\cite{guan2025fedpw}, and FedCE~\cite{jiang2023fedce} use update directions to set aggregation weights. FedPW also considers performance, while FedCE incorporates leave-one-client-out prediction errors. FedLD~\cite{zeng2024fedld} and the server variant of GC-Fed~\cite{seo2026gcfed} exploit direction-space structure through principal-gradient calibration and projection onto a fixed subspace, respectively.

Even when client updates agree on a descent direction, the shared model can remain sensitive to small parameter perturbations. Flatness optimization targets this sensitivity by seeking parameters with low loss in their neighborhood.

\subsection{Flatness optimization in federated learning}
\label{sec:related_flatness}
SAM~\cite{foret2021sam} seeks low loss in a parameter neighborhood, and ASAM~\cite{kwon2021asam} adapts that neighborhood to parameter scales. FedSAM~\cite{qu2022fedsam} and Caldarola et al.~\cite{caldarola2022flatminima} use local SAM, with the latter also studying ASAM and server-side weight averaging. DP-FedSAM~\cite{shi2025dpfedsam} applies local SAM under client-level differential privacy. As local flatness may not transfer to the aggregate, FedLESAM~\cite{fan2024fedlesam} and FedGMT~\cite{li2025fedgmt} use global model history to estimate perturbations with one backward pass per local step. MoFedSAM~\cite{qu2022fedsam} and FedNSAM~\cite{liu2025fednsam} use momentum to bring global information into local updates, with FedNSAM using it for perturbation and extrapolation. FedGF~\cite{lee2024fedgf} and FedGloSS~\cite{caldarola2025fedgloss} directly target global flatness through global perturbations in local training and server-side SAM, respectively.

Even with global guidance, client descent directions can still conflict, and clients can respond differently to the same perturbation. These distinct effects motivate joint designs that combine flatness optimization with update coordination.

\subsection{Joint update coordination and flatness optimization}
\label{sec:related_joint}
FedSpeed~\cite{sun2023fedspeed} and FedSOL~\cite{lee2024fedsol} combine local perturbation-based training with proximal information, using proximal correction to reduce update bias and proximal gradients to define perturbations, respectively. FedSMOO~\cite{sun2023fedsmoo}, FedGAMMA~\cite{dai2024fedgamma}, and FedTOGA~\cite{li2026fedtoga} link local consistency with global flatness via shared perturbations with dynamic regularization, control variates, and global-update guidance for perturbations and regularization, respectively. FedWMSAM~\cite{li2025fedwmsam} and FedScar~\cite{lu2026fedscar} use cross-round information, through server momentum and cosine-based weighting in FedWMSAM and accumulated geometric bias correction in FedScar. FedISM~\cite{wu2024fedism} pairs local flatness optimization with aggregation weights based on client sharpness estimates.

Although these methods combine update correction or aggregation with flatness, their coupling rules leave independent composition across method families unresolved. Our framework bridges this gap in multi-step federated training by retaining descent and flatness correction signals along one local trajectory and relating them to model updates. The server coordinates the branches with separate states before fusion, and explicit adapters let compatible mechanisms share the coordination interface.

%% file: sections/03_method.tex
\section{Method}
\label{sec:method}

\input{sections/03_framework_figure}

As shown in \cref{fig:framework}, the framework proceeds in three stages. Clients extract and accumulate two signals along a single local training trajectory. The server coordinates client directions within each branch, then fuses the two directions to update the shared model.

\subsection{Motivation for combining update coordination and flatness optimization}
\label{sec:overview}
Consider a federated system with $N$ clients and local data distributions $\mathcal D_i$. Each model contains shared parameters $\theta$ and optional private parameters $\phi_i$. Client $i$ minimizes
\begin{equation}
 L_i(\theta,\phi_i)=\E_{(x,y)\sim\mathcal D_i}
 \bigl[\ell(q_{\phi_i}(f_\theta(x)),y)\bigr],
 \label{eq:risk}
\end{equation}
where $f_\theta$ is the shared feature extractor and $q_{\phi_i}$ is the prediction head. A fully shared model includes the head in $\theta$ and omits $\phi_i$. At communication round $t$, participating clients $\mathcal I_t$ receive $\theta_t$ and perform local training.

To understand how one client's update affects another, we fix the private parameters and evaluate the clients at a common shared parameter $\theta$. Let $g_i=\nabla_\theta L_i$ and $H_j=\nabla_\theta^2 L_j$. For the update $\theta-\eta g_i$ proposed by client $i$, the change in client $j$'s loss is
\begin{equation}
 \begin{aligned}
 &L_j(\theta-\eta g_i)-L_j(\theta)\\
 &\quad=\underbrace{-\eta g_j^\top g_i}_{\text{first-order term}}
 +\underbrace{\frac{\eta^2}{2}g_i^\top H_jg_i}_{\text{second-order term}}+R_{ij}.
 \end{aligned}
 \label{eq:cross_client_taylor}
\end{equation}
We suppress the fixed private parameters. In a smooth neighborhood with a Lipschitz Hessian, the remainder satisfies $R_{ij}=\mathcal O(\eta^3\norm{g_i}^3)$.

The first-order term captures directional agreement across clients. If $g_j^\top g_i<0$, client $i$'s update increases client $j$'s loss to first order. A positive inner product contributes to descent. Update coordination addresses conflicts by adjusting client directions or their aggregation weights. For a coordinated direction $d$, increasing $g_j^\top d$ improves the first-order contribution to client $j$'s descent.

At a finite step size, positive directional curvature can offset this first-order benefit, while negative curvature can reinforce it. Flatness optimization uses gradients at perturbed parameters to seek low-loss regions that are less sensitive to parameter changes~\cite{foret2021sam}. In the quadratic model, the relevant quantity is the directional curvature $c_j(d)=d^\top H_jd/\norm{d}^2$. At a fixed step size and direction norm, reducing positive directional curvature reduces the second-order penalty.

The descent signal describes the client's update direction at the current parameters, while the flatness correction signal captures the change in its gradient under perturbation. The two signals describe different relationships across clients, and mixing them before coordination can obscure those relationships. To preserve the information within each branch, we record both signals along the same local trajectory and coordinate them independently at the server before fusion.

\subsection{Local signal extraction and accumulation}
\label{sec:trajectory}
Each client follows one local training trajectory while accumulating the descent and flatness correction signals separately. Let $\bar g_{i,t}$ and $\bar h_{i,t}$ denote the averages of these signals over client $i$'s $K_{i,t}$ local steps in round $t$. Under the local SGD conditions given below, the two averages exactly reconstruct the total local increment of the shared parameters:
\begin{equation}
 \Delta_{i,t}:=\theta_t-\theta_{i,K_{i,t}}
 =\eta_{\loc}K_{i,t}(\bar g_{i,t}+\bar h_{i,t}).
 \label{eq:reconstruction}
\end{equation}

At local step $k$, client $i$ has parameters $w_{i,k}=(\theta_{i,k},\phi_{i,k})$. The selected flatness optimization mechanism $\probe$ uses local, global, or historical information to construct perturbed parameters $\widetilde w_{i,k}$. Using a paired gradient decomposition~\cite{phan2025fdse}, we evaluate the ordinary and perturbed gradients on the shared coordinates with the same minibatch $B_{i,k}$ and stochastic state:
\begin{equation}
 \begin{aligned}
 g_{i,k}&=\nabla_\theta L_i(w_{i,k}\,;B_{i,k}),\\
 p_{i,k}&=\left.\nabla_\theta L_i(w\,;B_{i,k})\right|_{w=\widetilde w_{i,k}},\\
 h_{i,k}&=p_{i,k}-g_{i,k}.
 \end{aligned}
 \label{eq:branches}
\end{equation}
The perturbation is held fixed during differentiation. Let $\epsilon_{i,k}=\widetilde w_{i,k}-w_{i,k}$ and let $\mathsf H_{i,k}$ denote the minibatch Hessian with respect to all local parameters. Under a locally Lipschitz Hessian, a small perturbation gives
\begin{equation}
 h_{i,k}=[\mathsf H_{i,k}\epsilon_{i,k}]_\theta
 +\mathcal O(\norm{\epsilon_{i,k}}^2),
 \label{eq:probe_response}
\end{equation}
where $[\cdot]_\theta$ selects the shared coordinates. Thus, $g_{i,k}$ provides first-order descent information and $h_{i,k}$ captures the local geometric response to the perturbation.

Starting from $\theta_{i,0}=\theta_t$, the client updates the shared parameters using $g_{i,k}+h_{i,k}=p_{i,k}$ and updates any private parameters with the corresponding perturbed gradient:
\begin{equation}
 \theta_{i,k+1}=\theta_{i,k}-\eta_{\loc}(g_{i,k}+h_{i,k}).
 \label{eq:local}
\end{equation}
After local training, the client computes the two step averages separately:
\begin{equation}
 \begin{aligned}
 \bar g_{i,t}&=\frac{1}{K_{i,t}}\sum\nolimits_{k=0}^{K_{i,t}-1}g_{i,k},\\
 \text{and}\quad\bar h_{i,t}&=\frac{1}{K_{i,t}}\sum\nolimits_{k=0}^{K_{i,t}-1}h_{i,k}.
 \end{aligned}
 \label{eq:means}
\end{equation}
These averages are the yellow and red dashed uploads on the right of \cref{fig:framework}. The client also provides its local step count and any sample counts required by the coordinator. Objective-dependent coordinators additionally receive the ordinary-loss summary $\bar\ell_{i,t}$ and the signed perturbation-response summary $\bar s_{i,t}$. Both summaries are weighted by minibatch sample counts, with the response computed as $L_i(\widetilde w_{i,k}\,;B_{i,k})-L_i(w_{i,k}\,;B_{i,k})$.

With a constant local learning rate and zero momentum and weight decay on the shared parameters, summing \cref{eq:local} over the local trajectory and substituting \cref{eq:means} gives the endpoint identity in \cref{eq:reconstruction}. The identity recovers the shared parameter increment using the local learning rate and step count. Unequal local step counts can bias the effective federated objective~\cite{wang2020fednova}.

\subsection{Independent server coordination and update fusion}
\label{sec:fusion}
Consider what a coordinator sees when the descent and flatness correction signals are mixed before aggregation. Suppressing the round index and writing $\bar p_i=\bar g_i+\bar h_i$, the inner product between two mixed directions is
\begin{equation}
 \bar p_i^\top\bar p_j=\bar g_i^\top\bar g_j+\bar h_i^\top\bar h_j+\bar g_i^\top\bar h_j+\bar h_i^\top\bar g_j.
 \label{eq:mixed_relations}
\end{equation}
The first two terms describe client relationships within the descent and flatness branches. The remaining terms couple the two branches. For a coordinator that uses inner products to determine projections or weights, these cross terms can change the coordination decision. Separate coordination preserves the relationships within each branch.

Let $\coord$ denote the selected coordination operator. The server applies the same operator separately to the two branches, using their respective summaries and states:
\begin{equation}
 \begin{aligned}
 (d_t^g,\hist_{t+1}^g)
 &=\coord(\{(\bar g_{i,t},\bar\ell_{i,t})\}_{i\in\mathcal I_t},\hist_t^g),\\
 \text{and}\quad(d_t^h,\hist_{t+1}^h)
 &=\coord(\{(\bar h_{i,t},\bar s_{i,t})\}_{i\in\mathcal I_t},\hist_t^h).
 \end{aligned}
 \label{eq:coordination}
\end{equation}
The operator determines which summaries it requires. For a stateless operator, the corresponding state is empty. Otherwise, each branch maintains its own history, such as past projection directions or adaptive weights.

The coordinated directions $d_t^g$ and $d_t^h$ correspond to the yellow and red solid arrows in \cref{fig:framework}. The server fuses them and updates the shared model with optimizer $\update$:
\begin{equation}
 \begin{aligned}
 d_t&=d_t^g+\lambda d_t^h,\\
 \text{and}\quad(\theta_{t+1},\Omega_{t+1})&=\update(\theta_t,d_t,\Omega_t).
 \end{aligned}
 \label{eq:fuse}
\end{equation}
Here $\lambda$ controls the contribution of the flatness branch, and $\Omega_t$ is the server optimizer state. The black arrow in \cref{fig:framework} denotes the fused direction, and the outer path broadcasts the updated shared model for the next round.

For a fixed linear operator $\coord(\{u_i\})=\sum_i a_i u_i$ with identical weights across calls, the order of coordination and fusion does not change the resulting direction:
\begin{equation}
 \coord(\{\bar g_i+\lambda\bar h_i\})
 =\coord(\{\bar g_i\})+\lambda\coord(\{\bar h_i\}).
 \label{eq:linear}
\end{equation}
Input-dependent projection, normalization, weighting, or history can produce different directions. By preserving the relationships and states within each branch, the framework lets compatible flatness optimization mechanisms share a coordination interface and be selected independently of the coordination rule. Supplementary Sec.~C specifies the adapters. \Cref{alg:framework} summarizes the full procedure.

\begin{algorithm}[t]
\caption{Bridging update coordination and flatness}
\label{alg:framework}
\small
\begin{algorithmic}[1]
\Require Shared parameters $\theta_0$, mechanisms $\probe$, $\coord$, $\update$, and $\lambda$
\State Initialize client states, $\hist^g$, $\hist^h$, and $\Omega$
\State Initialize references required by $\probe$
\For{round $t=0,\ldots,T-1$}
 \State Select clients and broadcast $\theta_t$ and references
 \For{client $i\in\mathcal{I}_t$ in parallel}
  \State $\theta_{i,0}\gets\theta_t$, $C_i^g\gets0$, $C_i^h\gets0$
  \For{$k=0,\ldots,K_{i,t}-1$}
   \State Obtain $g_{i,k}$, $p_{i,k}$, $h_{i,k}$ using \cref{eq:branches}
   \State Update shared parameters with \cref{eq:local}
   \State Update any private parameters locally
   \State $C_i^g\gets C_i^g+g_{i,k}$, $C_i^h\gets C_i^h+h_{i,k}$
  \EndFor
  \State Upload $C_i^g/K_{i,t}$, $C_i^h/K_{i,t}$, and summaries
 \EndFor
 \State Coordinate the two groups independently using \cref{eq:coordination}
 \State Fuse directions and obtain $\theta_{t+1}$ using \cref{eq:fuse}
 \State Refresh perturbation references from client increments
\EndFor
\end{algorithmic}
\end{algorithm}

%% file: sections/03_framework_figure.tex
\begin{figure*}[t]
    \centering
    \includegraphics[width=\textwidth]{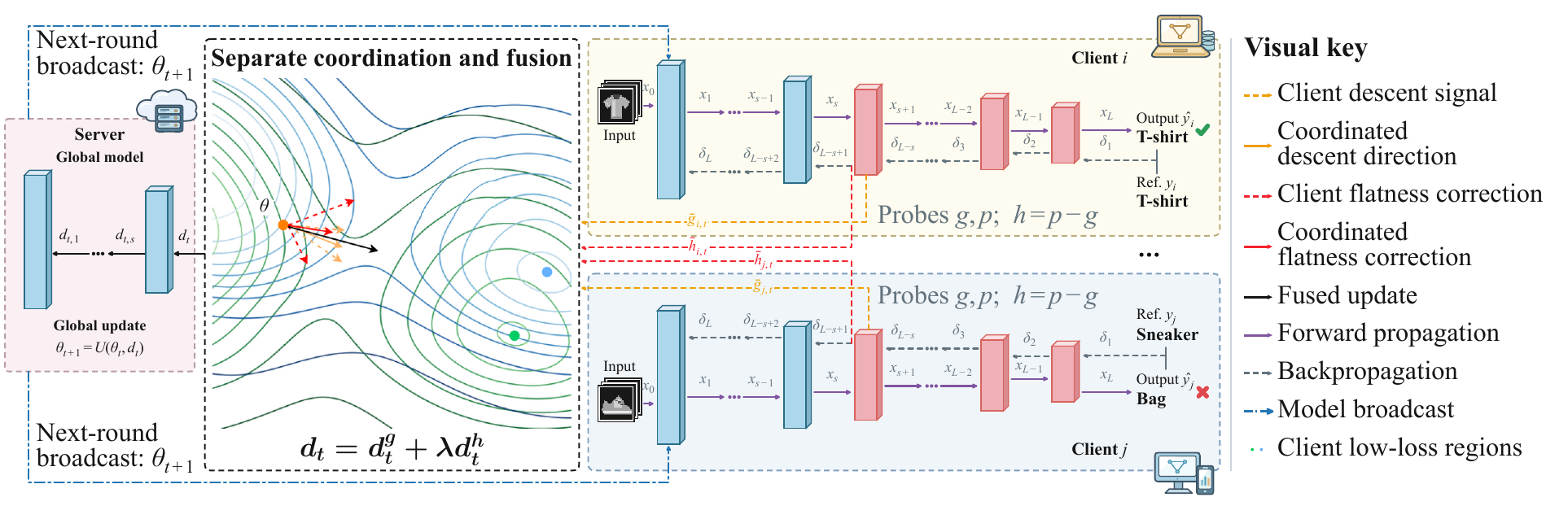}
    \caption{\textbf{An overview of the proposed framework.}
    Clients form $h=p-g$ from paired gradient evaluations and separately accumulate descent and flatness signals along the same local trajectory. The server coordinates the two groups, fuses $d_t=d_t^g+\lambda d_t^h$, and updates the shared model. Yellow and red dashed arrows carry client descent and flatness signals, respectively, while black arrows represent fused updates. Purple solid, gray dashed, and blue dash-dotted arrows show forward propagation, backpropagation, and model broadcast. Blue and pink layers denote shared and optional client-specific parameters. With $L$ layers and $s$ shared layers, $x_\ell$ denotes a forward feature, $\delta_r$ indexes backward signals from the output ($r=L-\ell+1$), and $d_{t,\ell}$ denotes the layer-$\ell$ block of $d_t$. The landscape and clothing icons are schematic.}
    \label{fig:framework}
\end{figure*}

%% file: sections/04_experiments.tex
\section{Experiments}
\label{sec:experiments}
We pair five update coordinators with four flatness mechanisms on Fashion-MNIST, CIFAR-100, Office-Home, and Retina, evaluating 80 combinations. A matched Fashion-MNIST experiment examines the effect of coordination order, and a perturbation sweep and cost analysis characterize the framework's sensitivity and training overhead. The four main result tables summarize runs with seeds 0, 1, and 2.

\subsection{Experimental setup}
\label{sec:protocol}
\paragraph{Datasets and client partitions.}
We evaluate the framework across four complementary federated settings, spanning label skew in a small client population, label heterogeneity with more clients, real-world domain shifts, and class imbalance in medical image classification. Fashion-MNIST~\cite{xiao2017fashion} uses five clients with two preferred classes each, and CIFAR-100~\cite{krizhevsky2009cifar} uses twenty clients with class-wise Dirichlet partitions at concentration $\alpha=0.5$. Office-Home~\cite{venkateswara2017officehome} treats its four visual domains as clients, with class-stratified splits within each domain. Retina uses the Normal/Diseased version of the Kaggle Diabetic Retinopathy dataset~\cite{yan2023retina} released with FedLD~\cite{zeng2024fedld}, using its five-client assignment with patient-group splits to evaluate heterogeneous class proportions and client sizes. Split ratios, preprocessing, and seed-dependent partition rules are specified in the supplementary material.

\paragraph{Compared methods.}
We evaluate methods from 2021--2025 by pairing five coordinators, FedFV~\cite{wang2021fedfv}, FedMGDA+~\cite{hu2022fedmgda}, FedMDFG-Agg~\cite{pan2023fedmdfg}, FedLD-Agg~\cite{zeng2024fedld}, and FedAWA-Agg~\cite{shi2025fedawa}, with FedSAM~\cite{qu2022fedsam}, FedASAM~\cite{caldarola2022flatminima}, FedGF~\cite{lee2024fedgf}, and FedGloSS~\cite{caldarola2025fedgloss}. To isolate the benefit of fusion, each combination is compared with its coordinator without flatness under matched architectures, partitions, budgets, and optimization settings. FedAvg~\cite{mcmahan2017fedavg} provides a common reference. Agg denotes the coordination component integrated into our training procedure.

\paragraph{Models and training.}
All compared methods use the same architecture per dataset, with a shared ResNet backbone~\cite{he2016resnet} and private client classifiers. We train ResNet-18 and ResNet-20 from scratch on Fashion-MNIST and CIFAR-100, respectively, and use ImageNet-pretrained ResNet-18 on Office-Home and Retina. Local shared parameters use momentum-free SGD, while private classifiers and coordinated server updates use Adam~\cite{kingma2015adam}. All clients participate each round, with learning rates and budgets in \Cref{tab:training_settings}. See the supplement for model and optimizer details.

\input{tables/experimental_setup}

\input{tables/fashion_main}
\input{tables/cifar_main}
\input{tables/officehome_main}
\input{tables/retina_main}

A data epoch is one pass over each client's training set. Per round, clients take up to five minibatch steps on Fashion-MNIST, CIFAR-100, and Office-Home and complete one local epoch on Retina. Partitions and initialization are matched across methods within each seed (0, 1, and 2).

\paragraph{Evaluation and model selection.}
For client-specific test accuracy $A_i$, we report
\begin{equation}
 \macroacc=\frac{1}{N}\sum_{i=1}^N A_i\;\text{and}\;
 \minacc=\min_i A_i.
 \label{eq:metrics}
\end{equation}
We select checkpoints by the highest client-mean validation accuracy and evaluate all metrics with their associated client states. Results are summarized as mean and sample standard deviation over three seeds. Accuracy gains are paired differences from the corresponding coordinator without flatness, reported in percentage points (pp).

\subsection{Combining coordination with flatness signals}
\label{sec:combinations}
\setlength{\intextsep}{8pt}
\begin{figure}[!htb]
    \centering
    \includegraphics[width=\columnwidth]{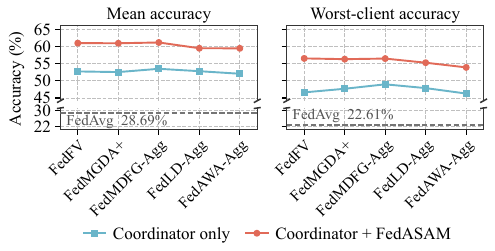}
    \captionsetup{skip=6pt}
    \caption{\textbf{Effect of adding FedASAM on CIFAR-100.} Points are test-accuracy means over three seeds.}
    \label{fig:cifar_asam_comparison}
\end{figure}

Across the four benchmarks, 76 of the 80 evaluated combinations improve client-macro test accuracy over their coordination-only baselines, and 74 of the 80 improve both macro and worst-client accuracy. These joint improvements show that the gains generally extend to the lower end of client performance. On Fashion-MNIST, all 20 combinations improve macro accuracy despite baseline accuracies near 95\%.

As shown in \cref{fig:cifar_asam_comparison}, FedASAM improves macro accuracy by 6.71--8.43 pp and worst-client accuracy by 7.43--9.94 pp across all five coordinators on label-heterogeneous CIFAR-100. The 8.43 pp macro gain is the largest across the four benchmarks. Across all four benchmarks, FedASAM improves both metrics over the corresponding coordination-only baselines in all 20 evaluated pairings.

Office-Home shows why independent component selection matters. With FedFV, FedGloSS improves macro accuracy by 4.50 pp compared with 3.93 pp for FedGF. With FedAWA-Agg, the ranking reverses, with gains of 8.17 pp for FedGF and 0.62 pp for FedGloSS. This reversal shows that the preferred flatness mechanism can depend on the coordinator even within the same benchmark. The framework defines the fusion rule once and uses adapters to connect compatible components to it. Researchers can therefore swap either component independently and evaluate alternative pairings under the same training procedure.

\subsection{Separate versus direct coordination}
\label{sec:direct}

\setlength{\textfloatsep}{8pt}
\setlength{\dbltextfloatsep}{8pt}
\setlength{\intextsep}{8pt}
\input{sections/04_branch_replay}
\input{sections/04_validation_trajectories}

\noindent\textbf{Same-input replay.} Both orders use identical descent and flatness signals in each of nine five-client replay blocks from Fashion-MNIST. In Fig.~\ref{fig:branch_replay}(a), cross terms flip 5 of 90 pairwise signs, resolving four conflicts and introducing one.

\noindent\textbf{Effect on the server direction.} With the same fixed averaging weights, both directions agree to numerical precision, as predicted by \cref{eq:linear}. In block 8, mixing preserves all ten pairwise signs, yet the FedFV directions differ by $1.22^\circ$. At equal step lengths, separate coordination predicts a 4.69\% smaller first-order loss decrease for client 4, as shown in Fig.~\ref{fig:branch_replay}(b).

\noindent\textbf{Full-training behavior.} On Fashion-MNIST with seed~2, we evaluate FedASAM with separate and direct server coordination, holding the perturbation rule, radius, parameter scope, local update rule, and server optimizer fixed. Both variants improve macro and worst-client test accuracy over the corresponding coordination-only baselines across all five coordinators, with macro accuracy differing by at most 0.25 pp. Separate coordination retains these gains with independent histories that support branch inspection and component replacement. For seed 2, direct FedASAM improves both validation metrics over FedAWA-Agg alone in the final 20 epochs on Fashion-MNIST and CIFAR-100 in \cref{fig:validation_trajectories}. In the top panel of \cref{fig:descent_conflicts}, FedFV with separately coordinated FedASAM has fewer descent conflicts throughout most of training. Over the final 20 Fashion-MNIST epochs, the conflict rate is 1.00\% with FedASAM and 24.78\% with FedFV alone.

\noindent\textbf{Perturbation sensitivity.} The bottom panel of \cref{fig:descent_conflicts} shows that adding FedASAM to FedAWA-Agg reduces the validation-loss increase at every nonzero radius tested. At 0.5\%, macro validation loss rises by 0.54 with FedASAM versus 1.43 with FedAWA-Agg alone.

\input{sections/04_descent_conflicts}

\subsection{Communication and computation}
\label{sec:cost}
The framework achieves the gains in \cref{sec:combinations} with matched round budgets and the original inference cost. Separate coordination doubles the uncompressed uplink per round to $2md$ scalars ($8md$ FP32 bytes) for $m$ clients and $d$ shared parameters. Both orders use the same downlink for a given mechanism. Separate coordination adds a server coordinator call, vector fusion, and operator-dependent history storage. Client overhead depends on whether the ordinary gradient is already available.

On Fashion-MNIST at seed~2, FedFV + FedASAM uses 86,000 backward passes over 1,800 rounds with either coordination order. Separate coordination improves worst-client test accuracy by 0.65 pp. Logged training and validation take 59.34 vs.\ 58.56 min, an increase of 1.32\%, with hardware and load unverified. Separate coordination reaches 96.0\% mean validation accuracy for three consecutive epochs about 9.3\% sooner, while direct coordination reaches 95.6\% and 95.8\% sooner. The extra uplink supports independent histories and component replacement under the same fusion rule.

%% file: tables/experimental_setup.tex
\par
\begingroup
\ifdefined\nolinenumbers
  \nolinenumbers
\fi
\begin{center}
\begin{minipage}{\linewidth}
\centering
\captionof{table}{Training settings and communication budgets.}
\label{tab:training_settings}
\footnotesize
\setlength{\tabcolsep}{2.7pt}
\renewcommand{\arraystretch}{1.15}
\begin{tabular}{@{}lrrrrcc@{}}
\toprule
Dataset & Clients & Batch & \shortstack{Data\\epochs} & Rounds & \shortstack{Local\\LR} & \shortstack{Server\\LR} \\
\midrule
Fashion-MNIST & 5 & 256 & 200 & 1,800 & $10^{-3}$ & $10^{-3}$ \\
CIFAR-100 & 20 & 128 & 300 & 1,500 & $10^{-3}$ & $10^{-3}$ \\
Office-Home & 4 & 16 & 100 & 3,900 & $10^{-3}$ & $10^{-4}$ \\
Retina & 5 & 50 & 100 & 100 & $10^{-2}$ & $10^{-4}$ \\
\bottomrule
\end{tabular}
\end{minipage}
\end{center}
\par
\endgroup

%% file: tables/fashion_main.tex
\begin{table*}[t]
\centering
\caption{Fashion-MNIST results across seeds 0, 1, and 2 at validation-selected checkpoints. Entries report mean $\pm$ sample standard deviation ($n=3$) for macro and worst-client test accuracy (\%), and paired macro gains $\Delta$ (pp) over the corresponding ordinary coordinator. Bold denotes a combination mean no lower than its ordinary baseline before rounding. FedAvg is a shared parameter-averaging reference.}
\label{tab:fashion}
\begingroup
\fontsize{5.8}{7}\selectfont
\setlength{\tabcolsep}{0.8pt}
\setlength{\arrayrulewidth}{0.3pt}
\renewcommand{\arraystretch}{1}
\newlength{\fashioncellwidth}
\setlength{\fashioncellwidth}{\dimexpr(\textwidth-32\tabcolsep-5\arrayrulewidth)/16\relax}
\newcommand{\fashionvenue}[1]{{\fontsize{4.4}{5.2}\selectfont\mbox{#1}}}
\newcommand{\fashionnumber}[1]{{\fontsize{6.2}{7.2}\selectfont\mbox{#1}}}
\newcommand{\fashionstat}[2]{\mbox{{\fontsize{6.2}{7.2}\selectfont #1}\kern0.5pt{\fontsize{4.2}{5}\selectfont\textpm #2}}}
\newcommand{\fashioncell}[1]{\parbox[c][16pt][c]{\fashioncellwidth}{\centering #1}}
\begin{tabular}{c|ccc|ccc|ccc|ccc|ccc}
\toprule
\resultspanhead{Flatness\\mechanism} & \multicolumn{3}{c|}{FedFV\,\fashionvenue{(IJCAI 2021)}} & \multicolumn{3}{c|}{FedMGDA+\,\fashionvenue{(IEEE TNSE 2022)}} & \multicolumn{3}{c|}{FedMDFG-Agg\,\fashionvenue{(AAAI 2023)}} & \multicolumn{3}{c|}{FedLD-Agg\,\fashionvenue{(MICCAI 2024)}} & \multicolumn{3}{c}{FedAWA-Agg\,\fashionvenue{(CVPR 2025)}} \\
\cmidrule(lr){2-4}\cmidrule(lr){5-7}\cmidrule(lr){8-10}\cmidrule(lr){11-13}\cmidrule(lr){14-16}
 & Mean$\uparrow$ & Worst$\uparrow$ & Gain$\uparrow$ & Mean$\uparrow$ & Worst$\uparrow$ & Gain$\uparrow$ & Mean$\uparrow$ & Worst$\uparrow$ & Gain$\uparrow$ & Mean$\uparrow$ & Worst$\uparrow$ & Gain$\uparrow$ & Mean$\uparrow$ & Worst$\uparrow$ & Gain$\uparrow$ \\
\midrule
\fashioncell{FedAvg\\\fashionvenue{(AISTATS 2017)}} & \multicolumn{15}{c}{\fashionnumber{Mean: \fashionstat{92.25}{0.24}\qquad Worst: \fashionstat{90.10}{0.18}}} \\
\rowcolor{black!7}
\fashioncell{Ordinary} & \fashioncell{\fashionnumber{\fashionstat{95.16}{0.29}}} & \fashioncell{\fashionnumber{\fashionstat{93.38}{0.69}}} & \fashioncell{\fashionnumber{---}} & \fashioncell{\fashionnumber{\fashionstat{94.91}{0.14}}} & \fashioncell{\fashionnumber{\fashionstat{93.35}{0.39}}} & \fashioncell{\fashionnumber{---}} & \fashioncell{\fashionnumber{\fashionstat{95.20}{0.11}}} & \fashioncell{\fashionnumber{\fashionstat{93.85}{0.66}}} & \fashioncell{\fashionnumber{---}} & \fashioncell{\fashionnumber{\fashionstat{95.00}{0.10}}} & \fashioncell{\fashionnumber{\fashionstat{93.65}{0.26}}} & \fashioncell{\fashionnumber{---}} & \fashioncell{\fashionnumber{\fashionstat{95.10}{0.12}}} & \fashioncell{\fashionnumber{\fashionstat{93.55}{0.13}}} & \fashioncell{\fashionnumber{---}} \\
\fashioncell{+ FedSAM\\\fashionvenue{(ICML 2022)}} & \fashioncell{\fashionnumber{\textbf{\fashionstat{95.62}{0.07}}}} & \fashioncell{\fashionnumber{\textbf{\fashionstat{94.40}{0.30}}}} & \fashioncell{\fashionnumber{\fashionstat{+0.46}{0.31}}} & \fashioncell{\fashionnumber{\textbf{\fashionstat{95.67}{0.04}}}} & \fashioncell{\fashionnumber{\textbf{\fashionstat{94.40}{0.30}}}} & \fashioncell{\fashionnumber{\fashionstat{+0.76}{0.11}}} & \fashioncell{\fashionnumber{\textbf{\fashionstat{95.73}{0.11}}}} & \fashioncell{\fashionnumber{\textbf{\fashionstat{94.47}{0.16}}}} & \fashioncell{\fashionnumber{\fashionstat{+0.53}{0.11}}} & \fashioncell{\fashionnumber{\textbf{\fashionstat{95.70}{0.14}}}} & \fashioncell{\fashionnumber{\textbf{\fashionstat{94.50}{0.18}}}} & \fashioncell{\fashionnumber{\fashionstat{+0.69}{0.09}}} & \fashioncell{\fashionnumber{\textbf{\fashionstat{95.63}{0.05}}}} & \fashioncell{\fashionnumber{\textbf{\fashionstat{94.27}{0.31}}}} & \fashioncell{\fashionnumber{\fashionstat{+0.53}{0.13}}} \\
\fashioncell{+ FedASAM\\\fashionvenue{(ECCV 2022)}} & \fashioncell{\fashionnumber{\textbf{\fashionstat{95.67}{0.14}}}} & \fashioncell{\fashionnumber{\textbf{\fashionstat{94.25}{0.13}}}} & \fashioncell{\fashionnumber{\fashionstat{+0.51}{0.42}}} & \fashioncell{\fashionnumber{\textbf{\fashionstat{95.68}{0.08}}}} & \fashioncell{\fashionnumber{\textbf{\fashionstat{94.43}{0.23}}}} & \fashioncell{\fashionnumber{\fashionstat{+0.77}{0.13}}} & \fashioncell{\fashionnumber{\textbf{\fashionstat{95.73}{0.06}}}} & \fashioncell{\fashionnumber{\textbf{\fashionstat{94.45}{0.39}}}} & \fashioncell{\fashionnumber{\fashionstat{+0.53}{0.10}}} & \fashioncell{\fashionnumber{\textbf{\fashionstat{95.65}{0.03}}}} & \fashioncell{\fashionnumber{\textbf{\fashionstat{94.52}{0.25}}}} & \fashioncell{\fashionnumber{\fashionstat{+0.65}{0.11}}} & \fashioncell{\fashionnumber{\textbf{\fashionstat{95.70}{0.11}}}} & \fashioncell{\fashionnumber{\textbf{\fashionstat{94.53}{0.10}}}} & \fashioncell{\fashionnumber{\fashionstat{+0.60}{0.09}}} \\
\fashioncell{+ FedGF\\\fashionvenue{(ICML 2024)}} & \fashioncell{\fashionnumber{\textbf{\fashionstat{95.59}{0.17}}}} & \fashioncell{\fashionnumber{\textbf{\fashionstat{94.28}{0.23}}}} & \fashioncell{\fashionnumber{\fashionstat{+0.43}{0.31}}} & \fashioncell{\fashionnumber{\textbf{\fashionstat{95.61}{0.23}}}} & \fashioncell{\fashionnumber{\textbf{\fashionstat{94.08}{0.54}}}} & \fashioncell{\fashionnumber{\fashionstat{+0.70}{0.18}}} & \fashioncell{\fashionnumber{\textbf{\fashionstat{95.65}{0.10}}}} & \fashioncell{\fashionnumber{\textbf{\fashionstat{94.20}{0.41}}}} & \fashioncell{\fashionnumber{\fashionstat{+0.45}{0.16}}} & \fashioncell{\fashionnumber{\textbf{\fashionstat{95.58}{0.03}}}} & \fashioncell{\fashionnumber{\textbf{\fashionstat{94.38}{0.26}}}} & \fashioncell{\fashionnumber{\fashionstat{+0.58}{0.07}}} & \fashioncell{\fashionnumber{\textbf{\fashionstat{95.67}{0.11}}}} & \fashioncell{\fashionnumber{\textbf{\fashionstat{94.28}{0.38}}}} & \fashioncell{\fashionnumber{\fashionstat{+0.57}{0.19}}} \\
\fashioncell{+ FedGloSS\\\fashionvenue{(CVPR 2025)}} & \fashioncell{\fashionnumber{\textbf{\fashionstat{95.44}{0.10}}}} & \fashioncell{\fashionnumber{\textbf{\fashionstat{94.23}{0.43}}}} & \fashioncell{\fashionnumber{\fashionstat{+0.28}{0.21}}} & \fashioncell{\fashionnumber{\textbf{\fashionstat{95.53}{0.08}}}} & \fashioncell{\fashionnumber{\textbf{\fashionstat{94.12}{0.08}}}} & \fashioncell{\fashionnumber{\fashionstat{+0.62}{0.19}}} & \fashioncell{\fashionnumber{\textbf{\fashionstat{95.46}{0.27}}}} & \fashioncell{\fashionnumber{\fashionstat{93.83}{0.60}}} & \fashioncell{\fashionnumber{\fashionstat{+0.26}{0.18}}} & \fashioncell{\fashionnumber{\textbf{\fashionstat{95.37}{0.27}}}} & \fashioncell{\fashionnumber{\textbf{\fashionstat{94.18}{0.19}}}} & \fashioncell{\fashionnumber{\fashionstat{+0.37}{0.21}}} & \fashioncell{\fashionnumber{\textbf{\fashionstat{95.40}{0.15}}}} & \fashioncell{\fashionnumber{\textbf{\fashionstat{94.07}{0.42}}}} & \fashioncell{\fashionnumber{\fashionstat{+0.30}{0.17}}} \\
\bottomrule
\end{tabular}
\typeout{FASHION-DATA-CELL-WIDTH=\the\fashioncellwidth}
\typeout{FASHION-DATA-CELL-HEIGHT=16pt}
\typeout{FASHION-MEAN-FONT=6.2pt; UNCERTAINTY-FONT=4.2pt; LABEL-FONT=5.8pt; VENUE-FONT=4.4pt}
\endgroup
\end{table*}

%% file: tables/cifar_main.tex
\begin{table*}[t]
\centering
\caption{CIFAR-100 results across seeds 0, 1, and 2 at validation-selected checkpoints. Entries report mean $\pm$ sample standard deviation ($n=3$) for macro and worst-client test accuracy (\%), and paired macro gains $\Delta$ (pp) over the corresponding ordinary coordinator. Bold denotes a combination mean no lower than its ordinary baseline before rounding. FedAvg is a shared parameter-averaging reference.}
\label{tab:cifar}
\begingroup
\fontsize{5.8}{7}\selectfont
\setlength{\tabcolsep}{0.8pt}
\setlength{\arrayrulewidth}{0.3pt}
\renewcommand{\arraystretch}{1}
\newlength{\cifarcellwidth}
\setlength{\cifarcellwidth}{\dimexpr(\textwidth-32\tabcolsep-5\arrayrulewidth)/16\relax}
\newcommand{\cifarvenue}[1]{{\fontsize{4.4}{5.2}\selectfont\mbox{#1}}}
\newcommand{\cifarnumber}[1]{{\fontsize{6.2}{7.2}\selectfont\mbox{#1}}}
\newcommand{\cifarstat}[2]{\mbox{{\fontsize{6.2}{7.2}\selectfont #1}\kern0.5pt{\fontsize{4.2}{5}\selectfont\textpm #2}}}
\newcommand{\cifarcell}[1]{\parbox[c][16pt][c]{\cifarcellwidth}{\centering #1}}
\begin{tabular}{c|ccc|ccc|ccc|ccc|ccc}
\toprule
\resultspanhead{Flatness\\mechanism} & \multicolumn{3}{c|}{FedFV\,\cifarvenue{(IJCAI 2021)}} & \multicolumn{3}{c|}{FedMGDA+\,\cifarvenue{(IEEE TNSE 2022)}} & \multicolumn{3}{c|}{FedMDFG-Agg\,\cifarvenue{(AAAI 2023)}} & \multicolumn{3}{c|}{FedLD-Agg\,\cifarvenue{(MICCAI 2024)}} & \multicolumn{3}{c}{FedAWA-Agg\,\cifarvenue{(CVPR 2025)}} \\
\cmidrule(lr){2-4}\cmidrule(lr){5-7}\cmidrule(lr){8-10}\cmidrule(lr){11-13}\cmidrule(lr){14-16}
 & Mean$\uparrow$ & Worst$\uparrow$ & Gain$\uparrow$ & Mean$\uparrow$ & Worst$\uparrow$ & Gain$\uparrow$ & Mean$\uparrow$ & Worst$\uparrow$ & Gain$\uparrow$ & Mean$\uparrow$ & Worst$\uparrow$ & Gain$\uparrow$ & Mean$\uparrow$ & Worst$\uparrow$ & Gain$\uparrow$ \\
\midrule
\cifarcell{FedAvg\\\cifarvenue{(AISTATS 2017)}} & \multicolumn{15}{c}{\cifarnumber{Mean: \cifarstat{28.69}{0.73}\qquad Worst: \cifarstat{22.61}{1.36}}} \\
\rowcolor{black!7}
\cifarcell{Ordinary} & \cifarcell{\cifarnumber{\cifarstat{52.78}{0.83}}} & \cifarcell{\cifarnumber{\cifarstat{46.65}{1.45}}} & \cifarcell{\cifarnumber{---}} & \cifarcell{\cifarnumber{\cifarstat{52.59}{0.62}}} & \cifarcell{\cifarnumber{\cifarstat{47.71}{1.65}}} & \cifarcell{\cifarnumber{---}} & \cifarcell{\cifarnumber{\cifarstat{53.55}{1.32}}} & \cifarcell{\cifarnumber{\cifarstat{49.01}{0.15}}} & \cifarcell{\cifarnumber{---}} & \cifarcell{\cifarnumber{\cifarstat{52.85}{1.15}}} & \cifarcell{\cifarnumber{\cifarstat{47.90}{1.67}}} & \cifarcell{\cifarnumber{---}} & \cifarcell{\cifarnumber{\cifarstat{52.10}{1.20}}} & \cifarcell{\cifarnumber{\cifarstat{46.31}{1.18}}} & \cifarcell{\cifarnumber{---}} \\
\cifarcell{+ FedSAM\\\cifarvenue{(ICML 2022)}} & \cifarcell{\cifarnumber{\textbf{\cifarstat{54.63}{1.04}}}} & \cifarcell{\cifarnumber{\textbf{\cifarstat{50.25}{2.77}}}} & \cifarcell{\cifarnumber{\cifarstat{+1.85}{0.47}}} & \cifarcell{\cifarnumber{\textbf{\cifarstat{57.15}{0.85}}}} & \cifarcell{\cifarnumber{\textbf{\cifarstat{52.47}{1.35}}}} & \cifarcell{\cifarnumber{\cifarstat{+4.57}{0.57}}} & \cifarcell{\cifarnumber{\textbf{\cifarstat{53.61}{1.47}}}} & \cifarcell{\cifarnumber{\textbf{\cifarstat{49.11}{1.71}}}} & \cifarcell{\cifarnumber{\cifarstat{+0.07}{0.17}}} & \cifarcell{\cifarnumber{\textbf{\cifarstat{55.24}{1.81}}}} & \cifarcell{\cifarnumber{\textbf{\cifarstat{50.94}{1.54}}}} & \cifarcell{\cifarnumber{\cifarstat{+2.38}{0.67}}} & \cifarcell{\cifarnumber{\textbf{\cifarstat{57.52}{0.89}}}} & \cifarcell{\cifarnumber{\textbf{\cifarstat{52.83}{0.80}}}} & \cifarcell{\cifarnumber{\cifarstat{+5.42}{0.73}}} \\
\cifarcell{+ FedASAM\\\cifarvenue{(ECCV 2022)}} & \cifarcell{\cifarnumber{\textbf{\cifarstat{61.06}{0.78}}}} & \cifarcell{\cifarnumber{\textbf{\cifarstat{56.59}{1.60}}}} & \cifarcell{\cifarnumber{\cifarstat{+8.28}{0.33}}} & \cifarcell{\cifarnumber{\textbf{\cifarstat{61.02}{1.21}}}} & \cifarcell{\cifarnumber{\textbf{\cifarstat{56.37}{1.37}}}} & \cifarcell{\cifarnumber{\cifarstat{+8.43}{0.60}}} & \cifarcell{\cifarnumber{\textbf{\cifarstat{61.25}{1.21}}}} & \cifarcell{\cifarnumber{\textbf{\cifarstat{56.54}{1.38}}}} & \cifarcell{\cifarnumber{\cifarstat{+7.70}{0.48}}} & \cifarcell{\cifarnumber{\textbf{\cifarstat{59.56}{1.12}}}} & \cifarcell{\cifarnumber{\textbf{\cifarstat{55.33}{1.07}}}} & \cifarcell{\cifarnumber{\cifarstat{+6.71}{0.22}}} & \cifarcell{\cifarnumber{\textbf{\cifarstat{59.54}{1.10}}}} & \cifarcell{\cifarnumber{\textbf{\cifarstat{53.96}{1.68}}}} & \cifarcell{\cifarnumber{\cifarstat{+7.44}{0.34}}} \\
\cifarcell{+ FedGF\\\cifarvenue{(ICML 2024)}} & \cifarcell{\cifarnumber{\textbf{\cifarstat{53.34}{0.90}}}} & \cifarcell{\cifarnumber{\textbf{\cifarstat{48.11}{1.63}}}} & \cifarcell{\cifarnumber{\cifarstat{+0.56}{0.45}}} & \cifarcell{\cifarnumber{\textbf{\cifarstat{54.89}{0.85}}}} & \cifarcell{\cifarnumber{\textbf{\cifarstat{50.46}{0.82}}}} & \cifarcell{\cifarnumber{\cifarstat{+2.30}{0.26}}} & \cifarcell{\cifarnumber{\textbf{\cifarstat{53.84}{0.53}}}} & \cifarcell{\cifarnumber{\textbf{\cifarstat{49.20}{0.50}}}} & \cifarcell{\cifarnumber{\cifarstat{+0.29}{1.15}}} & \cifarcell{\cifarnumber{\textbf{\cifarstat{53.01}{1.76}}}} & \cifarcell{\cifarnumber{\textbf{\cifarstat{48.55}{2.21}}}} & \cifarcell{\cifarnumber{\cifarstat{+0.15}{0.61}}} & \cifarcell{\cifarnumber{\textbf{\cifarstat{53.94}{1.18}}}} & \cifarcell{\cifarnumber{\textbf{\cifarstat{49.05}{1.96}}}} & \cifarcell{\cifarnumber{\cifarstat{+1.84}{0.09}}} \\
\cifarcell{+ FedGloSS\\\cifarvenue{(CVPR 2025)}} & \cifarcell{\cifarnumber{\textbf{\cifarstat{54.00}{1.47}}}} & \cifarcell{\cifarnumber{\textbf{\cifarstat{48.58}{2.23}}}} & \cifarcell{\cifarnumber{\cifarstat{+1.22}{0.86}}} & \cifarcell{\cifarnumber{\textbf{\cifarstat{53.46}{1.08}}}} & \cifarcell{\cifarnumber{\textbf{\cifarstat{48.37}{1.64}}}} & \cifarcell{\cifarnumber{\cifarstat{+0.87}{0.46}}} & \cifarcell{\cifarnumber{\textbf{\cifarstat{54.18}{1.65}}}} & \cifarcell{\cifarnumber{\textbf{\cifarstat{50.85}{1.37}}}} & \cifarcell{\cifarnumber{\cifarstat{+0.63}{0.49}}} & \cifarcell{\cifarnumber{\textbf{\cifarstat{53.25}{0.97}}}} & \cifarcell{\cifarnumber{\textbf{\cifarstat{48.38}{1.22}}}} & \cifarcell{\cifarnumber{\cifarstat{+0.40}{0.60}}} & \cifarcell{\cifarnumber{\textbf{\cifarstat{52.74}{1.23}}}} & \cifarcell{\cifarnumber{\textbf{\cifarstat{47.63}{1.41}}}} & \cifarcell{\cifarnumber{\cifarstat{+0.64}{0.52}}} \\
\bottomrule
\end{tabular}
\typeout{CIFAR-DATA-CELL-WIDTH=\the\cifarcellwidth}
\typeout{CIFAR-DATA-CELL-HEIGHT=16pt}
\typeout{CIFAR-MEAN-FONT=6.2pt; UNCERTAINTY-FONT=4.2pt; LABEL-FONT=5.8pt; VENUE-FONT=4.4pt}
\endgroup
\end{table*}

%% file: tables/officehome_main.tex
\begin{table*}[t]
\centering
\caption{Office-Home test results across seeds 0, 1, and 2. Entries report mean $\pm$ sample standard deviation for client-macro and worst-client accuracy (\%), and paired macro gains $\Delta$ (pp) over the corresponding ordinary coordinator. Bold indicates an accuracy at least as high as the ordinary baseline. FedAvg is a shared reference. FedGloSS uses a localized variant.}
\label{tab:officehome}
\begingroup
\fontsize{5.8}{7}\selectfont
\setlength{\tabcolsep}{0.8pt}
\setlength{\arrayrulewidth}{0.3pt}
\renewcommand{\arraystretch}{1}
\newlength{\officehomecellwidth}
\setlength{\officehomecellwidth}{\dimexpr(\textwidth-32\tabcolsep-5\arrayrulewidth)/16\relax}
\newcommand{\officehomevenue}[1]{{\fontsize{4.4}{5.2}\selectfont\mbox{#1}}}
\newcommand{\officehomenumber}[1]{{\fontsize{6.2}{7.2}\selectfont\mbox{#1}}}
\newcommand{\officehomestat}[2]{\mbox{{\fontsize{6.2}{7.2}\selectfont #1}\kern0.5pt{\fontsize{4.2}{5}\selectfont\textpm #2}}}
\newcommand{\officehomecell}[1]{\parbox[c][16pt][c]{\officehomecellwidth}{\centering #1}}
\begin{tabular}{c|ccc|ccc|ccc|ccc|ccc}
\toprule
\resultspanhead{Flatness\\mechanism} & \multicolumn{3}{c|}{FedFV\,\officehomevenue{(IJCAI 2021)}} & \multicolumn{3}{c|}{FedMGDA+\,\officehomevenue{(IEEE TNSE 2022)}} & \multicolumn{3}{c|}{FedMDFG-Agg\,\officehomevenue{(AAAI 2023)}} & \multicolumn{3}{c|}{FedLD-Agg\,\officehomevenue{(MICCAI 2024)}} & \multicolumn{3}{c}{FedAWA-Agg\,\officehomevenue{(CVPR 2025)}} \\
\cmidrule(lr){2-4}\cmidrule(lr){5-7}\cmidrule(lr){8-10}\cmidrule(lr){11-13}\cmidrule(lr){14-16}
 & Mean$\uparrow$ & Worst$\uparrow$ & Gain$\uparrow$ & Mean$\uparrow$ & Worst$\uparrow$ & Gain$\uparrow$ & Mean$\uparrow$ & Worst$\uparrow$ & Gain$\uparrow$ & Mean$\uparrow$ & Worst$\uparrow$ & Gain$\uparrow$ & Mean$\uparrow$ & Worst$\uparrow$ & Gain$\uparrow$ \\
\midrule
\officehomecell{FedAvg\\\officehomevenue{(AISTATS 2017)}} & \multicolumn{15}{c}{\officehomenumber{Mean: \officehomestat{75.07}{0.30}\qquad Worst: \officehomestat{65.15}{0.83}}} \\
\rowcolor{black!7}
\officehomecell{Ordinary} & \officehomecell{\officehomenumber{\officehomestat{76.57}{0.30}}} & \officehomecell{\officehomenumber{\officehomestat{66.53}{2.07}}} & \officehomecell{\officehomenumber{---}} & \officehomecell{\officehomenumber{\officehomestat{75.95}{0.92}}} & \officehomecell{\officehomenumber{\officehomestat{66.60}{2.63}}} & \officehomecell{\officehomenumber{---}} & \officehomecell{\officehomenumber{\officehomestat{75.65}{0.80}}} & \officehomecell{\officehomenumber{\officehomestat{65.50}{1.75}}} & \officehomecell{\officehomenumber{---}} & \officehomecell{\officehomenumber{\officehomestat{75.10}{0.74}}} & \officehomecell{\officehomenumber{\officehomestat{65.64}{0.72}}} & \officehomecell{\officehomenumber{---}} & \officehomecell{\officehomenumber{\officehomestat{71.76}{0.66}}} & \officehomecell{\officehomenumber{\officehomestat{62.54}{1.40}}} & \officehomecell{\officehomenumber{---}} \\
\officehomecell{+ FedSAM\\\officehomevenue{(ICML 2022)}} & \officehomecell{\officehomenumber{\textbf{\officehomestat{79.98}{0.27}}}} & \officehomecell{\officehomenumber{\textbf{\officehomestat{70.31}{0.74}}}} & \officehomecell{\officehomenumber{\officehomestat{+3.41}{0.55}}} & \officehomecell{\officehomenumber{\textbf{\officehomestat{77.77}{0.33}}}} & \officehomecell{\officehomenumber{\textbf{\officehomestat{67.22}{0.55}}}} & \officehomecell{\officehomenumber{\officehomestat{+1.81}{0.73}}} & \officehomecell{\officehomenumber{\textbf{\officehomestat{79.45}{0.18}}}} & \officehomecell{\officehomenumber{\textbf{\officehomestat{69.21}{0.52}}}} & \officehomecell{\officehomenumber{\officehomestat{+3.80}{0.71}}} & \officehomecell{\officehomenumber{\textbf{\officehomestat{75.90}{0.90}}}} & \officehomecell{\officehomenumber{\textbf{\officehomestat{65.91}{1.40}}}} & \officehomecell{\officehomenumber{\officehomestat{+0.80}{0.39}}} & \officehomecell{\officehomenumber{\textbf{\officehomestat{72.61}{0.15}}}} & \officehomecell{\officehomenumber{\textbf{\officehomestat{63.71}{0.90}}}} & \officehomecell{\officehomenumber{\officehomestat{+0.85}{0.79}}} \\
\officehomecell{+ FedASAM\\\officehomevenue{(ECCV 2022)}} & \officehomecell{\officehomenumber{\textbf{\officehomestat{79.97}{0.07}}}} & \officehomecell{\officehomenumber{\textbf{\officehomestat{69.76}{0.72}}}} & \officehomecell{\officehomenumber{\officehomestat{+3.40}{0.28}}} & \officehomecell{\officehomenumber{\textbf{\officehomestat{80.88}{0.23}}}} & \officehomecell{\officehomenumber{\textbf{\officehomestat{72.03}{1.14}}}} & \officehomecell{\officehomenumber{\officehomestat{+4.93}{0.25}}} & \officehomecell{\officehomenumber{\textbf{\officehomestat{79.92}{0.39}}}} & \officehomecell{\officehomenumber{\textbf{\officehomestat{69.48}{0.71}}}} & \officehomecell{\officehomenumber{\officehomestat{+4.27}{0.49}}} & \officehomecell{\officehomenumber{\textbf{\officehomestat{79.40}{0.78}}}} & \officehomecell{\officehomenumber{\textbf{\officehomestat{70.03}{2.00}}}} & \officehomecell{\officehomenumber{\officehomestat{+4.29}{0.72}}} & \officehomecell{\officehomenumber{\textbf{\officehomestat{74.13}{0.73}}}} & \officehomecell{\officehomenumber{\textbf{\officehomestat{64.26}{1.75}}}} & \officehomecell{\officehomenumber{\officehomestat{+2.37}{0.26}}} \\
\officehomecell{+ FedGF\\\officehomevenue{(ICML 2024)}} & \officehomecell{\officehomenumber{\textbf{\officehomestat{80.50}{0.49}}}} & \officehomecell{\officehomenumber{\textbf{\officehomestat{72.03}{0.52}}}} & \officehomecell{\officehomenumber{\officehomestat{+3.93}{0.23}}} & \officehomecell{\officehomenumber{\textbf{\officehomestat{80.22}{0.86}}}} & \officehomecell{\officehomenumber{\textbf{\officehomestat{70.52}{1.24}}}} & \officehomecell{\officehomenumber{\officehomestat{+4.26}{0.47}}} & \officehomecell{\officehomenumber{\textbf{\officehomestat{77.56}{0.69}}}} & \officehomecell{\officehomenumber{\textbf{\officehomestat{67.84}{1.64}}}} & \officehomecell{\officehomenumber{\officehomestat{+1.91}{1.19}}} & \officehomecell{\officehomenumber{\textbf{\officehomestat{79.77}{0.14}}}} & \officehomecell{\officehomenumber{\textbf{\officehomestat{70.52}{1.29}}}} & \officehomecell{\officehomenumber{\officehomestat{+4.66}{0.60}}} & \officehomecell{\officehomenumber{\textbf{\officehomestat{79.94}{0.30}}}} & \officehomecell{\officehomenumber{\textbf{\officehomestat{71.89}{0.93}}}} & \officehomecell{\officehomenumber{\officehomestat{+8.17}{0.76}}} \\
\officehomecell{+ FedGloSS\\\officehomevenue{(CVPR 2025)}} & \officehomecell{\officehomenumber{\textbf{\officehomestat{81.07}{0.27}}}} & \officehomecell{\officehomenumber{\textbf{\officehomestat{72.71}{0.72}}}} & \officehomecell{\officehomenumber{\officehomestat{+4.50}{0.09}}} & \officehomecell{\officehomenumber{\textbf{\officehomestat{80.59}{0.53}}}} & \officehomecell{\officehomenumber{\textbf{\officehomestat{71.96}{1.35}}}} & \officehomecell{\officehomenumber{\officehomestat{+4.64}{0.33}}} & \officehomecell{\officehomenumber{\textbf{\officehomestat{77.56}{0.29}}}} & \officehomecell{\officehomenumber{\textbf{\officehomestat{67.29}{1.78}}}} & \officehomecell{\officehomenumber{\officehomestat{+1.91}{0.91}}} & \officehomecell{\officehomenumber{\textbf{\officehomestat{75.89}{0.26}}}} & \officehomecell{\officehomenumber{\textbf{\officehomestat{65.84}{1.02}}}} & \officehomecell{\officehomenumber{\officehomestat{+0.78}{0.62}}} & \officehomecell{\officehomenumber{\textbf{\officehomestat{72.38}{0.75}}}} & \officehomecell{\officehomenumber{\textbf{\officehomestat{62.89}{0.21}}}} & \officehomecell{\officehomenumber{\officehomestat{+0.62}{1.31}}} \\
\bottomrule
\end{tabular}
\endgroup
\end{table*}

%% file: tables/retina_main.tex
\begin{table*}[t]
\centering
\caption{Retina test results at validation-selected checkpoints. Entries report mean $\pm$ sample standard deviation for macro and worst-client accuracy (\%), and reported macro gains $\Delta$ (pp) over the corresponding ordinary coordinator. Bold denotes a combination mean no lower than its ordinary baseline. FedAvg is shown once as a shared reference. FedGloSS uses a localized variant.}
\label{tab:retina}
\begingroup
\fontsize{5.8}{7}\selectfont
\setlength{\tabcolsep}{0.8pt}
\setlength{\arrayrulewidth}{0.3pt}
\renewcommand{\arraystretch}{1}
\newlength{\retinacellwidth}
\setlength{\retinacellwidth}{\dimexpr(\textwidth-32\tabcolsep-5\arrayrulewidth)/16\relax}
\newcommand{\retinavenue}[1]{{\fontsize{4.4}{5.2}\selectfont\mbox{#1}}}
\newcommand{\retinanumber}[1]{{\fontsize{6.2}{7.2}\selectfont\mbox{#1}}}
\newcommand{\retinastat}[2]{\mbox{{\fontsize{6.2}{7.2}\selectfont #1}\kern0.5pt{\fontsize{4.2}{5}\selectfont\textpm #2}}}
\newcommand{\retinacell}[1]{\parbox[c][16pt][c]{\retinacellwidth}{\centering #1}}
\begin{tabular}{c|ccc|ccc|ccc|ccc|ccc}
\toprule
\resultspanhead{Flatness\\mechanism} & \multicolumn{3}{c|}{FedFV\,\retinavenue{(IJCAI 2021)}} & \multicolumn{3}{c|}{FedMGDA+\,\retinavenue{(IEEE TNSE 2022)}} & \multicolumn{3}{c|}{FedMDFG-Agg\,\retinavenue{(AAAI 2023)}} & \multicolumn{3}{c|}{FedLD-Agg\,\retinavenue{(MICCAI 2024)}} & \multicolumn{3}{c}{FedAWA-Agg\,\retinavenue{(CVPR 2025)}} \\
\cmidrule(lr){2-4}\cmidrule(lr){5-7}\cmidrule(lr){8-10}\cmidrule(lr){11-13}\cmidrule(lr){14-16}
 & Mean$\uparrow$ & Worst$\uparrow$ & Gain$\uparrow$ & Mean$\uparrow$ & Worst$\uparrow$ & Gain$\uparrow$ & Mean$\uparrow$ & Worst$\uparrow$ & Gain$\uparrow$ & Mean$\uparrow$ & Worst$\uparrow$ & Gain$\uparrow$ & Mean$\uparrow$ & Worst$\uparrow$ & Gain$\uparrow$ \\
\midrule
\retinacell{FedAvg\\\retinavenue{(AISTATS 2017)}} & \multicolumn{15}{c}{\retinanumber{Mean: \retinastat{77.40}{0.28}\qquad Worst: \retinastat{68.42}{0.43}}} \\
\rowcolor{black!7}
\retinacell{Ordinary} & \retinacell{\retinanumber{\retinastat{78.50}{1.81}}} & \retinacell{\retinanumber{\retinastat{70.29}{4.34}}} & \retinacell{\retinanumber{---}} & \retinacell{\retinanumber{\retinastat{79.27}{2.58}}} & \retinacell{\retinanumber{\retinastat{70.68}{5.78}}} & \retinacell{\retinanumber{---}} & \retinacell{\retinanumber{\retinastat{78.90}{0.20}}} & \retinacell{\retinanumber{\retinastat{71.06}{0.49}}} & \retinacell{\retinanumber{---}} & \retinacell{\retinanumber{\retinastat{77.78}{1.40}}} & \retinacell{\retinanumber{\retinastat{67.71}{3.10}}} & \retinacell{\retinanumber{---}} & \retinacell{\retinanumber{\retinastat{78.15}{1.22}}} & \retinacell{\retinanumber{\retinastat{70.14}{0.66}}} & \retinacell{\retinanumber{---}} \\
\retinacell{+ FedSAM\\\retinavenue{(ICML 2022)}} & \retinacell{\retinanumber{\retinastat{78.47}{1.32}}} & \retinacell{\retinanumber{\retinastat{70.07}{1.47}}} & \retinacell{\retinanumber{\retinastat{-0.04}{2.91}}} & \retinacell{\retinanumber{\textbf{\retinastat{81.83}{1.25}}}} & \retinacell{\retinanumber{\textbf{\retinastat{75.39}{1.23}}}} & \retinacell{\retinanumber{\retinastat{+2.56}{2.48}}} & \retinacell{\retinanumber{\retinastat{78.14}{1.80}}} & \retinacell{\retinanumber{\retinastat{68.46}{2.34}}} & \retinacell{\retinanumber{\retinastat{-0.76}{1.99}}} & \retinacell{\retinanumber{\retinastat{76.90}{2.62}}} & \retinacell{\retinanumber{\textbf{\retinastat{68.06}{3.16}}}} & \retinacell{\retinanumber{\retinastat{-0.88}{1.83}}} & \retinacell{\retinanumber{\textbf{\retinastat{78.66}{1.08}}}} & \retinacell{\retinanumber{\textbf{\retinastat{70.19}{1.51}}}} & \retinacell{\retinanumber{\retinastat{+0.52}{2.28}}} \\
\retinacell{+ FedASAM\\\retinavenue{(ECCV 2022)}} & \retinacell{\retinanumber{\textbf{\retinastat{81.44}{0.82}}}} & \retinacell{\retinanumber{\textbf{\retinastat{74.74}{0.39}}}} & \retinacell{\retinanumber{\retinastat{+2.94}{2.30}}} & \retinacell{\retinanumber{\textbf{\retinastat{82.30}{1.39}}}} & \retinacell{\retinanumber{\textbf{\retinastat{75.88}{1.22}}}} & \retinacell{\retinanumber{\retinastat{+3.03}{3.94}}} & \retinacell{\retinanumber{\textbf{\retinastat{80.26}{1.52}}}} & \retinacell{\retinanumber{\textbf{\retinastat{72.76}{2.72}}}} & \retinacell{\retinanumber{\retinastat{+1.36}{1.70}}} & \retinacell{\retinanumber{\textbf{\retinastat{79.43}{1.31}}}} & \retinacell{\retinanumber{\textbf{\retinastat{72.01}{2.01}}}} & \retinacell{\retinanumber{\retinastat{+1.65}{0.10}}} & \retinacell{\retinanumber{\textbf{\retinastat{81.55}{1.03}}}} & \retinacell{\retinanumber{\textbf{\retinastat{75.33}{0.98}}}} & \retinacell{\retinanumber{\retinastat{+3.41}{1.93}}} \\
\retinacell{+ FedGF\\\retinavenue{(ICML 2024)}} & \retinacell{\retinanumber{\textbf{\retinastat{80.31}{0.65}}}} & \retinacell{\retinanumber{\textbf{\retinastat{73.70}{0.56}}}} & \retinacell{\retinanumber{\retinastat{+1.81}{2.46}}} & \retinacell{\retinanumber{\textbf{\retinastat{79.68}{1.65}}}} & \retinacell{\retinanumber{\textbf{\retinastat{70.94}{1.08}}}} & \retinacell{\retinanumber{\retinastat{+0.40}{4.00}}} & \retinacell{\retinanumber{\retinastat{76.94}{1.46}}} & \retinacell{\retinanumber{\retinastat{68.85}{2.54}}} & \retinacell{\retinanumber{\retinastat{-1.96}{1.60}}} & \retinacell{\retinanumber{\textbf{\retinastat{80.17}{0.60}}}} & \retinacell{\retinanumber{\textbf{\retinastat{73.89}{0.44}}}} & \retinacell{\retinanumber{\retinastat{+2.39}{0.96}}} & \retinacell{\retinanumber{\textbf{\retinastat{80.74}{0.47}}}} & \retinacell{\retinanumber{\textbf{\retinastat{74.42}{1.06}}}} & \retinacell{\retinanumber{\retinastat{+2.60}{1.68}}} \\
\retinacell{+ FedGloSS\\\retinavenue{(CVPR 2025)}} & \retinacell{\retinanumber{\textbf{\retinastat{79.38}{1.00}}}} & \retinacell{\retinanumber{\textbf{\retinastat{71.31}{0.81}}}} & \retinacell{\retinanumber{\retinastat{+0.88}{2.52}}} & \retinacell{\retinanumber{\textbf{\retinastat{80.11}{1.31}}}} & \retinacell{\retinanumber{\textbf{\retinastat{71.09}{2.88}}}} & \retinacell{\retinanumber{\retinastat{+0.84}{1.36}}} & \retinacell{\retinanumber{\textbf{\retinastat{80.08}{0.15}}}} & \retinacell{\retinanumber{\textbf{\retinastat{71.45}{2.46}}}} & \retinacell{\retinanumber{\retinastat{+1.18}{0.23}}} & \retinacell{\retinanumber{\textbf{\retinastat{78.24}{1.53}}}} & \retinacell{\retinanumber{\textbf{\retinastat{70.85}{2.05}}}} & \retinacell{\retinanumber{\retinastat{+0.46}{2.33}}} & \retinacell{\retinanumber{\textbf{\retinastat{79.55}{0.31}}}} & \retinacell{\retinanumber{\retinastat{70.11}{1.45}}} & \retinacell{\retinanumber{\retinastat{+1.41}{0.90}}} \\
\bottomrule
\end{tabular}
\typeout{RETINA-DATA-CELL-WIDTH=\the\retinacellwidth}
\typeout{RETINA-DATA-CELL-HEIGHT=16pt}
\typeout{RETINA-MEAN-FONT=6.2pt; UNCERTAINTY-FONT=4.2pt; LABEL-FONT=5.8pt; VENUE-FONT=4.4pt}
\endgroup
\end{table*}

%% file: sections/04_branch_replay.tex
\begin{figure}[!htb]
    \centering
    \includegraphics[width=\columnwidth]{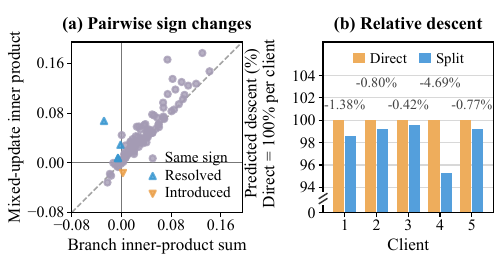}
    \caption{\textbf{Branch mixing and the server update.} We replay nine Fashion-MNIST A3 blocks from the seed-2 checkpoint at epoch 200. (a) uses a common scale for pairwise inner products. (b) compares predicted loss decreases at equal FedFV step lengths in block 8. Direct is 100\% per client. Labels give relative changes.}
    \label{fig:branch_replay}
\end{figure}

%% file: sections/04_validation_trajectories.tex
\begin{figure*}[!t]
    \centering
    \includegraphics[width=0.88\textwidth]{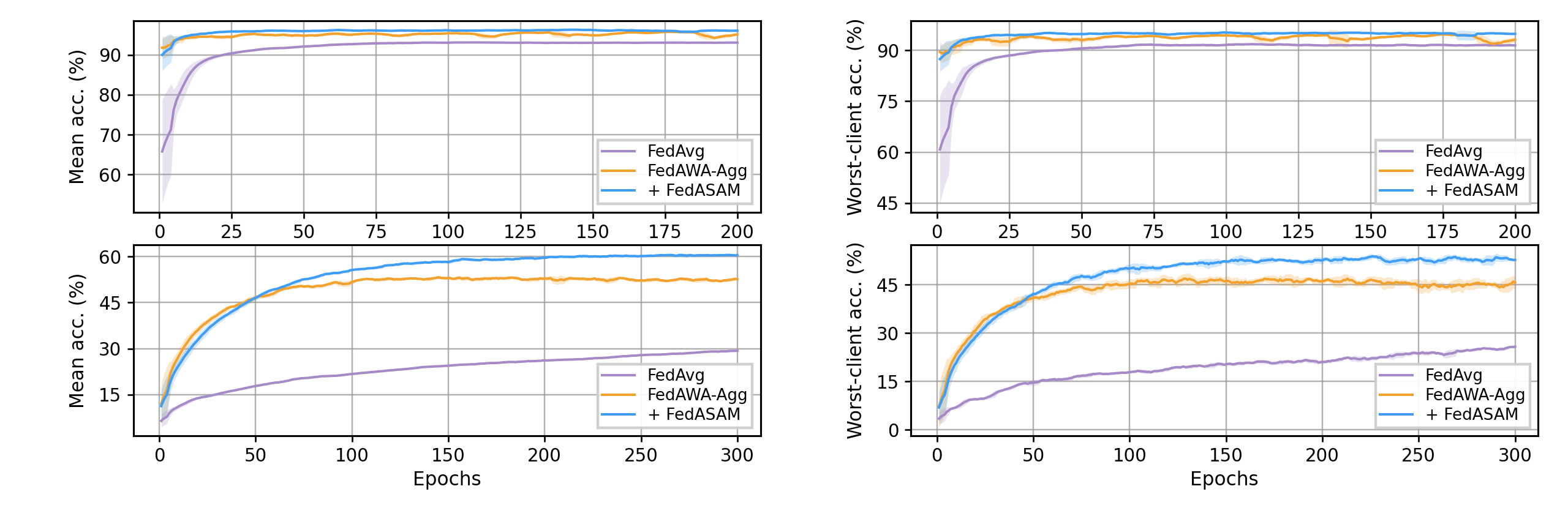}
    \caption{\textbf{Validation accuracy for seed 2.} Rows show Fashion-MNIST and CIFAR-100. Columns show client-macro and worst-client accuracy. FedASAM uses direct coordination of the perturbed gradient. Curves and bands show centered seven-epoch means and $\pm$ one sample standard deviation.}
    \label{fig:validation_trajectories}
\end{figure*}

%% file: sections/04_descent_conflicts.tex
\begin{figure}[!ht]
    \centering
    \includegraphics[width=\columnwidth,trim=145bp 0bp 163bp 48bp,clip]{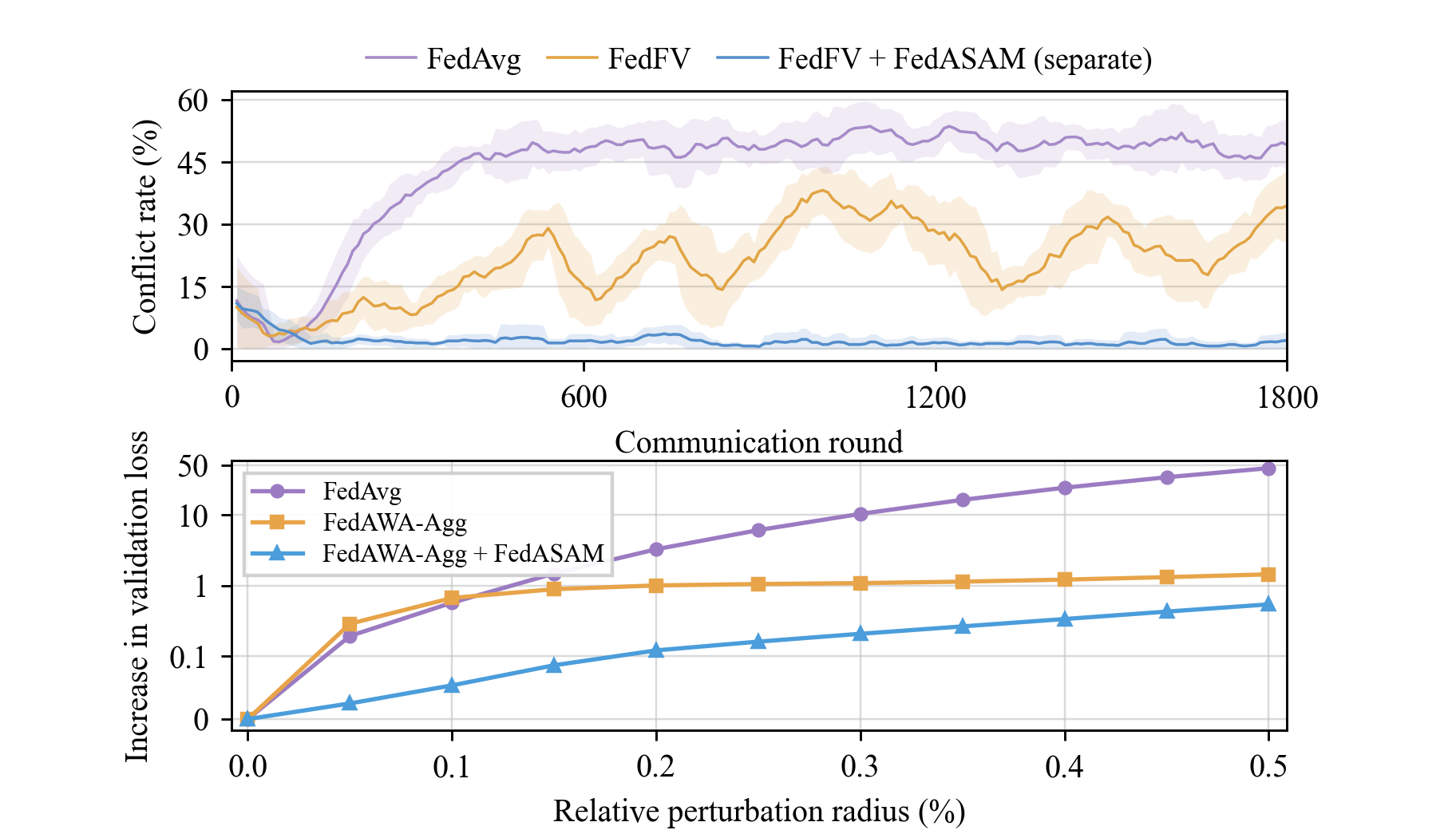}
    \caption{\textbf{Fashion-MNIST diagnostics for seed 2.} Top shows the fraction of client pairs with negative descent cosine before aggregation. Lines and bands show centered nine-epoch means and $\pm$ one sample standard deviation. Bottom shows the macro validation-loss increase under shared-backbone gradient ascent at the epoch-200 checkpoints. Radii are relative to the shared parameter norm. The loss axis uses a symmetric log scale.}
    \label{fig:descent_conflicts}
\end{figure}

%% file: sections/05_conclusion.tex
\section{Conclusion}
We proposed a unified federated learning framework that bridges update coordination and flatness optimization. First-order descent and second-order curvature effects explain their complementary roles, while endpoint reconstruction and coordination-order analysis establish consistency between the framework's trajectory signals, model updates, and coordination rules. Retaining both trajectory signals and isolating coordination states enables compatible components to be replaced independently. Experiments demonstrate improvements over ordinary-update baselines across diverse pairings while retaining the original inference architecture.

Without compression, separate coordination doubles the amount of data clients send to the server each round. In future work, we will compress the descent and flatness signals separately before upload to reduce transmission volume, while aiming to preserve independent coordination and the accuracy gains from combining the two mechanisms.

%% file: sections/supplementary_content.tex
\section{Local geometry and algebraic properties}
\label{sec:sup_geometry}
This supplement gives the assumptions behind the local analysis, the complete training procedure, adapter constants, and expanded results. Symbols follow the main paper. Shared parameters are denoted by $\theta$, private classifier parameters by $\phi_i$, and a shared descent displacement by $v$. Coordinator states are $\hist^g$ and $\hist^h$, distinct from Hessians denoted by $H$.

\subsection{Finite-step expansion}
Fix the private parameters, minibatch, and evaluation state, and write the resulting shared-parameter loss as $f(\theta)$. Suppose $f$ is twice differentiable on a neighborhood containing the segment from $\theta$ to $\theta-v$, and its Hessian is $L_H$-Lipschitz there. Taylor's theorem with an integral remainder gives
\begin{equation}
 f(\theta-v)-f(\theta)
 =-\nabla f(\theta)^\top v+\tfrac12v^\top H(\theta)v+R(v),
\end{equation}
with
\begin{equation}
 |R(v)|\leq\frac{L_H}{6}\norm{v}^3.
\end{equation}
To obtain the bound, write the second-order integral term as
\begin{equation}
 \int_0^1(1-s)v^\top H(\theta-sv)v\,ds.
\end{equation}
Subtracting $\tfrac12v^\top H(\theta)v$ and using the Hessian Lipschitz bound yields $L_H\norm{v}^3\int_0^1s(1-s)ds$. The same statement applies within a smooth region of a piecewise-smooth model. An activation-boundary crossing requires separate treatment of the local approximation.

For $v=\eta u$, define the directional curvature and the actual quadratic contribution by
\begin{equation}
 c_i(u)=\frac{u^\top H_i^{\shared\shared}u}{\norm{u}^2},\qquad
 Q_i(u,\eta)=\frac{\eta^2}{2}c_i(u)\norm{u}^2.
\end{equation}
The first quantity normalizes out the direction's magnitude. The second preserves both the magnitude and the step length. This distinction matters when a coordinator restores a norm or the server applies coordinate-wise preconditioning.

\begin{lemma}[Directional agreement and a finite step]
Nonnegative first-order agreement is insufficient by itself to determine the sign of a finite loss change.
\end{lemma}
\begin{proof}
Consider $f(x)=ax+\tfrac12bx^2$ at $x=0$, where $a>0$ and $b>0$. For a descent displacement $v>0$, the first-order term is $-av<0$, but the exact change is $-av+\tfrac12bv^2$. This change is positive whenever $v>2a/b$.
\end{proof}
This local property motivates retaining geometric response together with ordinary descent in the framework.

\subsection{Block response under a probe}
Let $w=(\theta,\phi)$ and let $\delta=(\delta_{\shared},\delta_\phi)$ be a fixed probe displacement for the gradient evaluation. With a locally Lipschitz Hessian,
\begin{equation}
 \nabla L(w+\delta)-\nabla L(w)
 =\int_0^1 H(w+s\delta)\delta\,ds.
\end{equation}
Taking shared coordinates gives
\begin{equation}
 h=H^{\shared\shared}(w)\delta_{\shared}
 +H^{\shared\phi}(w)\delta_\phi+r_{\shared},
\end{equation}
where $\norm{r_{\shared}}\leq(L_H/2)\norm{\delta}^2$. Joint backbone/head perturbation includes the cross block. A shared-only mechanism sets $\delta_\phi=0$. For FedGF, the effective displacement also includes the interpolation between the current local model and the round's shared starting model. The Taylor response uses that complete displacement.

The derivative in this construction is evaluated at the probe parameters with the generator detached. If a generator depends on $w$ or history, differentiating through its construction defines another update. We use the detached probe consistently across local and globally guided mechanisms.

\subsection{Trajectory reconstruction and coordination order}
For plain shared-parameter SGD with constant step size, zero momentum, and zero weight decay,
\begin{equation}
 \begin{split}
 \theta_t-\theta_{i,K_i}
 &=\sum_{k=0}^{K_i-1}(\theta_{i,k}-\theta_{i,k+1})\\
 &=\eta_{\loc}\sum_{k=0}^{K_i-1}(g_{i,k}+h_{i,k})\\
 &=\eta_{\loc}K_i(\bar g_i+\bar h_i).
 \end{split}
\end{equation}
This expression remains valid when the head changes during the local trajectory, because $p_{i,k}=g_{i,k}+h_{i,k}$ is defined at the actual state of that step. The averaged vectors summarize the trajectory. They need not equal gradients evaluated at the round's initial parameter vector.

For a fixed linear coordinator, both orders of composition equal $\sum_i a_i\bar g_i+\lambda\sum_i a_i\bar h_i$. With input-dependent coordination, the interaction is already visible in the pairwise inner product
\begin{equation}
 \begin{split}
 (g_i+\lambda h_i)^\top(g_j+\lambda h_j)
 ={}&g_i^\top g_j+\lambda g_i^\top h_j\\
 &+\lambda h_i^\top g_j+\lambda^2h_i^\top h_j.
 \end{split}
\end{equation}
Direct coordination sees these cross terms jointly. Separate coordination acts on the two within-branch relations and combines the outputs. The equality for linear averaging assumes identical fixed weights in each call, which is distinct from learning separate weights from each branch.

\section{Training procedure and state ownership}
\label{sec:sup_algorithm}
The main paper summarizes the training algorithm. Here we specify the client states, paired evaluations, and optimizer choices used in the reported experiments. The ordinary-update baseline uses the ordinary gradient for local shared and private updates and executes one coordinator. The direct variant uses the same perturbed local trajectory as the separate variant and coordinates the uploaded mean perturbed gradient once.

\paragraph{Parameter ownership and local optimization.}
The experiments use a shared representation $f_\theta$ and a private classifier $q_{\phi_i}$ for each client. BN running statistics remain local, while trainable BN affine parameters belong to $\theta$. The full ordinary and perturbed gradients are $G_{i,k}=\nabla_w L_i(w_{i,k}\,;B_{i,k})$ and $P_{i,k}=\left.\nabla_w L_i(w\,;B_{i,k})\right|_{w=\widetilde w_{i,k}}$. Their shared coordinates define $g_{i,k}$ and $p_{i,k}$. The shared local update uses plain SGD, and the private head uses persistent Adam with $(P_{i,k})_\phi$. After local training, the head, BN state, and private optimizer state remain at the client.

\paragraph{Server optimization.}
All coordination adapters use Adam~\cite{kingma2015adam} on $d_t=d_t^g+\lambda d_t^h$, with $\lambda=1$ except for CIFAR-100 FedMGDA+ with FedGloSS ($\lambda=0.5$). The server moments are formed from this fused direction and are separate from the two coordination histories. The actual parameter displacement is $v_t=\theta_t-\theta_{t+1}$, which is the displacement used in the finite-step loss analysis. Population sample weights, adaptive coordination weights, and client-macro evaluation weights have distinct roles.

\paragraph{Optimization settings.}
Local shared-parameter SGD has zero momentum and weight decay. Private classifiers use Adam at $10^{-3}$, and all learning rates remain constant. The main-paper budget table gives the local shared-parameter and server learning rates. Its server rates apply to the coordination methods using Adam, while FedAvg uses training-sample-weighted model averaging. Retina's local rate of $10^{-2}$ follows the FedLD reference configuration. Mechanism-specific hyperparameters are selected using validation performance.

\paragraph{Paired evaluations.}
The two passes use an identical input tensor, including any sampled augmentation. We snapshot the random state before the ordinary pass and replay it before the probe pass. Afterward, the random state resumes from the end of the ordinary pass. The ordinary pass supplies the retained BN running-state update. The probe pass uses training-mode batch statistics, but its extra running-state accumulation is discarded. Parameters are restored by copying saved tensors, which avoids subtraction-induced restoration drift.

\paragraph{Loss summaries.}
Let $\ell_{i,k}=L_i(w_{i,k}\,;B_{i,k})$ and $s_{i,k}=L_i(\widetilde w_{i,k}\,;B_{i,k})-L_i(w_{i,k}\,;B_{i,k})$. If minibatch $k$ has $b_{i,k}$ examples, the summaries are
\begin{equation}
 \bar\ell_i=\frac{\sum_k b_{i,k}\ell_{i,k}}{\sum_k b_{i,k}},\qquad
 \bar s_i=\frac{\sum_k b_{i,k}s_{i,k}}{\sum_k b_{i,k}}.
\end{equation}
These summaries serve ordering or objective-dependent guidance. Each vector mean uses the step count $K_i$ instead. A response with negative value remains signed, which preserves the generator's observed behavior.

\paragraph{Independent persistent states.}
The two coordinator states $\hist^g$ and $\hist^h$ separately store any projection history, expected objective, weight logits, or spectral history. Private classifier Adam states persist locally across rounds. The server Adam state operates only on the fused direction. Probe-generator references and the FedGF indicator window form another state. A checkpoint associates the shared model with all corresponding client states and optimizer histories.

\paragraph{Unequal intervals and global reference.}
Whenever a component requires a model increment, it uses $\Delta_i=\eta_{\loc}K_i(\bar g_i+\bar h_i)$. The global mechanism reference is the fitting-sample-weighted average of these increments. FedAWA-Agg uses a branch-specific pseudo-endpoint $\theta_t-\eta_{\loc}K_i u_i$ in its weight objective, with the correct $K_i$ for each client. Its final direction remains a weighted mean of the branch's trajectory-average gradients.

\paragraph{Historical probes.}
The FedLESAM definition uses the model last received by a particular client. Under the evaluated full-participation setting, that is the previous round's shared model. Under partial participation, the reference is indexed by the client's own most recent participation. The framework supports such client-specific references, while the evaluated adapter set uses full participation. In particular, adaptive weighting and fairness-guidance implementations must preserve their client-index mappings when extended to a subset.

\paragraph{Identity probes.}
If $\widetilde w=w$, paired deterministic evaluations yield $h=0$. The implementation treats zero-radius probing as disabled and returns this identity case, including for globally interpolated mechanisms. In finite precision, the endpoint identity is evaluated with a small numerical tolerance. These identities give consistency checks independently of predictive performance.

\section{Component adaptations and settings}
\label{sec:sup_adapters}
This section specifies how the selected coordination and perturbation components are used in the common framework. Each coordinator operates on the shared-coordinate vectors of one branch, with its corresponding summaries and independent state.
The Agg suffix denotes coordination or aggregation components integrated into this training procedure. FedFV and FedMGDA+ retain the names of their coordination rules.

\subsection{Coordination components}
\paragraph{FedFV.}
The basic conflict projection replaces a vector $u_i$ by
\begin{equation}
 u_i-\frac{u_i^\top u_j}{\norm{u_j}^2}u_j
 \quad\text{when }u_i^\top u_j<0.
\end{equation}
Our adapter retains the source ordering, protected fraction $0.1$, historical window $10$, and mean-norm restoration~\cite{wang2021fedfv}. Ordinary-loss summaries order the descent branch, and signed response summaries order the correction branch.

\paragraph{FedMGDA+.}
For nonzero inputs, the adapter uses $\widehat u_i=u_i/\norm{u_i}$ and the constrained minimum-norm problem~\cite{hu2022fedmgda}
\begin{equation}
 \begin{split}
 a^*&=\argmin_{a\in\mathcal C}\norm{\sum_i a_i\widehat u_i}^{2},\\
 \mathcal C&=\{a\geq0:\mathbf{1}^\top a=1,\norm{a-a^0}_\infty\leq\varepsilon\},
 \end{split}
\end{equation}
where $a^0$ is the reference weight vector. We use $\varepsilon=1$ and the source numerical recovery path. Explicit zero-vector handling preserves the intended problem without introducing an additional regularizer. Normalization is performed separately in each branch.

\paragraph{FedMDFG-Agg.}
The adapter uses the fairness angle $11.25^\circ$, normalized directions, the fairness-guided quadratic program, and output norm adjustment~\cite{pan2023fedmdfg}. Each branch maintains its own expected-objective history. The source line-search step is replaced by the common server Adam update. The branch objective is $\bar\ell_i$ for descent and the signed response $\bar s_i$ for the flatness correction.

\paragraph{FedAWA-Agg.}
For each branch vector $u_i$, we form the pseudo-endpoint
\begin{equation}
 \xi_{i,t}=\theta_t-\eta_{\loc}K_{i,t}u_i
\end{equation}
and supply it to the client-vector weight objective~\cite{shi2025fedawa}. The adapter returns $\sum_i a_i u_i$. We use one weight-optimization epoch per round, Adam for weight logits, $\gamma=1$, and cosine distance. The current round uses the weights available before that round's weight update, and the newly optimized weights are used in the next round. Each branch has its own weight state, separate from global-model Adam state.

\paragraph{FedLD-Agg.}
The principal-direction aggregation is applied separately to each shared parameter tensor~\cite{zeng2024fedld}. Its eigendecomposition acts on the client-update Gram matrix. We retain the source sign rule, spectral weighting, scale construction, principal-axis ratio $0.8$, and history decay $0.99$. The two branches maintain their own decompositions and histories. Local training uses the classification loss described in the main paper, without the source method's margin regularizer.

\subsection{Perturbation components}
\paragraph{Local SAM and ASAM.}
For the full local parameter vector $w$ and ordinary gradient $G$, the selected local mechanisms use
\begin{equation}
 \epsilon_{\mathrm{SAM}}=\rho\frac{G}{\norm{G}+\epsilon_0},\qquad
 \epsilon_{\mathrm{ASAM}}=\rho\frac{D^2G}{\norm{DG}+\epsilon_0},
\end{equation}
with $D=\diag(|w|)$ and $\widetilde w=w+\epsilon$~\cite{foret2021sam,kwon2021asam}. The joint gradient norm includes the shared backbone and private head, with $\epsilon_0=10^{-12}$. ASAM has no additional parameter-scale offset, and its numerator is the elementwise product $w^2\odot G$. The reported ASAM radius 2 and SAM radius 0.5 therefore describe different perturbation metrics.

\paragraph{Global reference.}
The globally guided mechanisms perturb shared parameters only. Define $\nu(u)=u/\norm{u}$ for nonzero $u$ and $\nu(0)=0$. Their raw reference increment is
\begin{equation}
 r_t=\sum_{i\in\mathcal{S}_t}\frac{n_i}{\sum_{j\in\mathcal{S}_t}n_j}\Delta_{i,t},
\end{equation}
where $n_i$ is the number of fitting samples and $\Delta_{i,t}=\theta_t-\theta_{i,K_{i,t}}$. This reference is formed before coordination and server Adam, with a zero initial reference.

\paragraph{FedGF.}
Following FedGF~\cite{lee2024fedgf}, we evaluate the shared gradient at
\begin{equation}
 \begin{split}
 \widetilde\theta_{i,k}={}&(1-c_t)\big(\theta_{i,k}+\rho\nu(g_{i,k})\big)\\
 &+c_t\big(\theta_t+\rho\nu(r_{t-1})\big).
 \end{split}
\end{equation}
The coefficient $c_t$ is the mean of the available most recent $W=30$ indicators $\mathbf{1}[D_\tau>T_D]$, where $D_t=|\mathcal{S}_t|^{-1}\sum_{i\in\mathcal{S}_t}\norm{\Delta_{i,t}}$. We set $c_0=0$. The effective probe displacement includes the difference between the local and global model references as well as their perturbations.

For each dataset, $T_D$ is the median raw endpoint-norm statistic from the first data epoch of a seed-2 FedFV/FedGF pilot. This uses nine rounds on Fashion-MNIST and five on CIFAR-100. The thresholds are
\begin{align}
 T_D^{\mathrm{Fashion}}&=0.006881969980895519,\\
 T_D^{\mathrm{CIFAR}}&=0.0038716522161848844.
\end{align}
Every coordinator within a dataset uses the same value. Calibration uses training increments and does not access validation or test accuracy.

\paragraph{FedLESAM and FedGloSS.}
Let $\tau_i(t)$ be the round when client $i$ last received a global model. The historical mechanism from FedLESAM~\cite{fan2024fedlesam} and the global mechanism from FedGloSS~\cite{caldarola2025fedgloss} use, respectively,
\begin{equation}
 \epsilon_{i,t}^{\mathrm{LE}}=\rho\nu(\theta_{\tau_i(t)}-\theta_t),\qquad
 \epsilon_t^{\mathrm{GL}}=\rho\nu(r_{t-1}).
\end{equation}
Both evaluate gradients at $\widetilde\theta_{i,k}=\theta_{i,k}+\epsilon$ using a joint norm over shared coordinates. The FedLESAM shift is zero at the client's first participation. We retain unshifted local coordinates for the paired ordinary gradient. These adapters use the source perturbation constructions, with coordination and the global-model update supplied by the common framework. The FedGloSS adapter therefore has no ADMM dual state.

\section{Dataset and evaluation details}
\label{sec:sup_protocol}
All clients within a dataset predict the same classes. Different local heads express parameter ownership and client distributions, rather than different task definitions. Under a selected shared checkpoint, predictions use the associated head and BN state for each client.

\input{tables/fashion_full}
\input{tables/cifar_full}

\paragraph{Partition seeds and data traversal.}
Seeds 0, 1, and 2 control data splitting, random initialization, sample order, and augmentation. CIFAR-100 independently samples its Dirichlet allocation for each seed. Office-Home domain assignments and Retina client assignments remain fixed, while their internal training, validation, and test splits vary with the seed. Compared methods share partitions and initialization within each seed. All clients participate in each round. A complete data epoch visits every client's fitting examples once. For Fashion-MNIST, CIFAR-100, and Office-Home, this traversal is distributed across rounds with at most five minibatch updates per client per round, retaining the final incomplete batch. Retina performs one complete local epoch per round.

\paragraph{Fashion-MNIST.}
The five clients each have two preferred classes. For every class, 80\% of its samples belong to its corresponding client and 5\% to each remaining client, yielding equal total client sizes with different class proportions. A ten-percent validation holdout from the official training set gives 54,000 fitting and 6,000 validation examples, alongside the 10,000 official test examples. The same class-allocation proportions apply across all three splits, producing 10,800 fitting, 1,200 validation, and 2,000 test examples per client. Grayscale inputs are bilinearly resized to $36\times36$, scaled by $1/255$, and receive no random image augmentation. The ResNet-18 is trained from scratch with a single-channel first convolution and a shared $512\to100$ projection followed by ReLU. Its private linear classifier maps these features to ten classes. Evaluation batch size is 512. The fitting batches produce nine communication rounds per data epoch.

\paragraph{CIFAR-100.}
Twenty clients receive class-wise Dirichlet allocations with $\alpha=0.5$. A ten-percent holdout within every training class gives 45,000 fitting and 5,000 validation examples, with 10,000 official test examples. The same class-allocation proportions apply across fitting, validation, and test assignments. Training augmentation uses four-pixel zero padding, random cropping, and horizontal flipping. Both normalization mean and standard deviation are 0.5, and evaluation preprocessing is deterministic. The ResNet-20 is trained from scratch with stage widths 16, 32, and 64, three blocks per stage, and 64 pooled features. Each private $64\to100$ linear head contains 6,500 parameters and has persistent Adam state. Each seed uses five communication rounds per complete data epoch. Local intervals contain two to five batches across these partitions, covering each fitting example once. All 300-epoch results use the same total data exposure. Runs extended from an earlier checkpoint preserve optimizer, local, coordinator, mechanism, and random states.

\paragraph{Office-Home.}
The four client domains are Art, Clipart, Product, and Real-World~\cite{venkateswara2017officehome}. Within each domain and class, seeded shuffling precedes the 70\%/10\%/20\% split. Validation and test counts are rounded within each class, and the remaining examples form the fitting set. Training resizes the shorter image side to 256 pixels, applies a random $224\times224$ crop, and flips horizontally with probability 0.5. Evaluation uses a center crop. The ImageNet-pretrained ResNet-18 backbone remains trainable and supplies 512 features to each private 65-class classifier. At batch size 16, the clients' complete data traversal spans 39 communication rounds, giving 3,900 rounds over 100 epochs.

\paragraph{Retina.}
The processed Normal/Diseased data originate from the Kaggle Diabetic Retinopathy dataset~\cite{yan2023retina} and retain the five-client Split-2 ownership released with FedLD~\cite{zeng2024fedld}. Approximate 70\%/10\%/20\% fitting, validation, and client-test splits are constructed by patient group. Images from the same patient and duplicate-image groups remain in the same subset across clients, preserving differences in client sizes and class proportions. Training applies a random resized crop to $256\times256$ with area fraction $[0.6,1.0]$ and aspect ratio $[0.75,4/3]$, random rotation within $\pm10$ degrees, and horizontal flipping with probability 0.5. Evaluation uses deterministic resizing and center cropping. Office-Home and Retina use ImageNet mean $(0.485,0.456,0.406)$ and standard deviation $(0.229,0.224,0.225)$. Retina uses a trainable pretrained ResNet-18 with private $512\to2$ linear heads. Each round visits every client's fitting examples once at batch size 50. Client-test AUROC uses the predicted Diseased probability and is averaged over clients containing both labels. Eligibility depends only on the test labels and is shared by all compared methods within a seed. The eligible client count accompanies the results. Accuracy retains all five clients.

\paragraph{Selection and metric aggregation.}
We select the checkpoint with the highest client-mean validation accuracy, resolving ties by the earliest evaluation. Accuracy and AUROC use its shared model and associated private heads and BN states. Client test accuracies are aggregated with equal weights, and the minimum is taken over clients at that same checkpoint. Metrics are first aggregated within each seed and then summarized as the mean and sample standard deviation over three seeds. Accuracy is reported in percent and AUROC on $[0,1]$. Accuracy gains are paired differences from the corresponding coordinator without flatness within each seed, then aggregated across seeds in percentage points. Tables report selected-checkpoint results throughout.

\paragraph{Numerical environment.}
The recorded training environment is Ubuntu 22.04 with an NVIDIA A100-PCIE-40GB GPU, Python 3.10.8, PyTorch 2.1.2+cu118, torchvision 0.16.2+cu118, CUDA runtime 11.8, and cuDNN 8.7.0. Numerical dependencies include NumPy 1.26.4, SciPy 1.11.4, CVXOPT 1.3.2, and quadprog 0.1.13. Training uses FP32 with AMP and TF32 disabled, deterministic algorithms, and cuDNN benchmarking disabled. Torch and OpenBLAS use four and one threads, respectively.

\section{Communication and computation details}
\label{sec:sup_cost}
The vector-payload accounting excludes scalar summaries and metadata and assumes uncompressed FP32. For $m$ participating clients with $d$ shared parameters, a single-vector upload uses $md$ scalars or $4md$ bytes, while dual-branch uploads use $2md$ scalars or $8md$ bytes. Both branches are sent within the same communication round. Downlink includes the shared model and any reference vectors required by the selected mechanism. Counting delivery to each client, each extra shared $d$-dimensional reference adds $4md$ bytes and accompanies the model in the same round.

For nonzero perturbations, paired gradient extraction performs one forward and backward pass at each of the ordinary and perturbed parameter settings. SAM/ASAM already computes both gradients, and FedGF reuses the ordinary gradient needed for its local perturbation. Constructing $h=p-g$ and accumulating the branches therefore introduces no extra network propagation for these mechanisms. FedGloSS constructs its perturbation from an existing global reference, so retaining the ordinary gradient adds one forward and backward pass relative to perturbed-only evaluation.

Relative to direct coordination, the server performs one extra coordinator call and vector fusion, without additional client-network propagation. Each branch retains the selected coordinator's own history, whose storage depends on the operator and its history length. The server optimizer acts once on the fused direction, so its state is not duplicated.

\section{Expanded results}
\label{sec:sup_results}
Both main result tables report means and sample standard deviations across seeds 0, 1, and 2. \Cref{tab:sup_fashion} retains seed-2 detail, including the additional FedLESAM configurations. \Cref{tab:sup_cifar} preserves earlier CIFAR-100 seed-2 reference runs with the probe settings in its caption, separately from the updated three-seed main results. Across-seed variation is distinct from variation among clients. Paired gains are computed before rounding, so subtracting two displayed means can differ by 0.01 pp from the reported gain.

For Fashion-MNIST, \Cref{tab:sup_fashion} reports the seed-2 comparison of FedASAM under direct and separate coordination. Across all five coordinators, both variants improve macro and worst-client test accuracy over the corresponding ordinary baselines, and their macro accuracies differ by at most 0.25 pp. The earlier CIFAR-100 reference runs include selected epochs. The updated main table reports all five coordinators paired with FedSAM, FedASAM, FedGF, and FedGloSS across three seeds.

Paired comparisons measure the benefit of adding a flatness mechanism with the shared/private parameter split and server optimizer held fixed. Both main tables report across-seed variation, while the supplementary tables provide individual-run detail. The specified parameter scopes and radii document the probe configurations used in these comparisons.

%% file: tables/fashion_full.tex
\begin{table*}[t]
\centering
\caption{Additional Fashion-MNIST seed-2 detail, supplementing the three-seed summary in the main paper. Each entry gives macro / worst-client test accuracy (\%). All methods use the same 200-epoch budget and validation selection. Both FedASAM variants use joint probes with $\rho=2$. The global probes use shared coordinates and $\rho=0.05$. The FedAvg reference is 92.15 / 90.05.}
\label{tab:sup_fashion}
\small\setlength{\tabcolsep}{3pt}
\begin{tabular}{@{}lcccccc@{}}
\toprule Coordinator & Ordinary & FedASAM (direct) & FedASAM (separate) & FedGF & FedLESAM & FedGloSS \\\midrule
FedFV & 95.36 / 93.85 & 95.62 / 94.10 & 95.68 / 94.75 & 95.76 / 94.50 & 95.27 / 93.95 & 95.27 / 94.20 \\
FedMGDA+ & 95.02 / 93.60 & 95.77 / 94.70 & 95.70 / 94.25 & 95.88 / 94.70 & 94.90 / 93.80 & 95.22 / 93.50 \\
FedMDFG-Agg & 95.32 / 94.10 & 95.78 / 94.70 & 95.83 / 94.65 & 95.59 / 94.10 & 95.30 / 93.95 & 95.24 / 93.85 \\
FedAWA-Agg & 95.19 / 93.65 & 95.83 / 94.50 & 95.58 / 94.20 & 95.72 / 94.65 & 95.21 / 93.25 & 95.36 / 93.70 \\
FedLD-Agg & 95.11 / 93.95 & 95.63 / 94.40 & 95.85 / 94.45 & 95.61 / 94.10 & 94.92 / 93.65 & 95.14 / 93.50 \\
\bottomrule\end{tabular}
\end{table*}

%% file: tables/cifar_full.tex
\begin{table*}[t]
\centering
\caption{Earlier CIFAR-100 seed-2 reference runs, retained separately from the updated three-seed main results. All combinations use separate coordination and 300 data epochs. Macro and worst-client accuracy are percentages, with gains in pp from unrounded records. The epoch column identifies the validation-selected checkpoint. These runs use SAM with $\rho=0.5$ and global probes with $\rho=0.05$. The FedAvg reference is 29.16 / 23.13, selected at epoch 299.}
\label{tab:sup_cifar}
\small\setlength{\tabcolsep}{10pt}
\begin{tabular}{@{}llrrrr@{}}
\toprule Coordinator & Mechanism & Macro & Worst & Gain & Epoch \\\midrule
FedFV & Ordinary & 53.68 & 47.16 &  & 176 \\
FedFV & SAM & 53.00 & 48.61 & -0.68 & 299 \\
FedFV & FedGF & 54.37 & 46.67 & +0.69 & 255 \\
FedFV & FedGloSS & 54.74 & 50.86 & +1.07 & 276 \\
\addlinespace[3pt]
FedMGDA+ & Ordinary & 53.27 & 47.65 &  & 147 \\
FedMGDA+ & SAM & 51.50 & 46.42 & -1.76 & 295 \\
\addlinespace[3pt]
FedMDFG-Agg & Ordinary & 54.98 & 49.14 &  & 237 \\
FedMDFG-Agg & SAM & 55.23 & 50.93 & +0.26 & 296 \\
FedMDFG-Agg & FedGF & 54.31 & 49.65 & -0.66 & 298 \\
FedMDFG-Agg & FedGloSS & 54.21 & 48.40 & -0.77 & 261 \\
\addlinespace[3pt]
FedAWA-Agg & Ordinary & 53.16 & 46.91 &  & 142 \\
FedAWA-Agg & SAM & 57.75 & 52.83 & +4.59 & 298 \\
FedAWA-Agg & FedGF & 53.11 & 49.31 & -0.04 & 285 \\
FedAWA-Agg & FedGloSS & 52.76 & 48.70 & -0.40 & 282 \\
\addlinespace[3pt]
FedLD-Agg & Ordinary & 53.97 & 49.77 &  & 239 \\
FedLD-Agg & SAM & 56.88 & 52.65 & +2.91 & 291 \\
FedLD-Agg & FedGF & 53.67 & 49.16 & -0.31 & 296 \\
FedLD-Agg & FedGloSS & 53.34 & 46.67 & -0.64 & 259 \\
\addlinespace[3pt]
\bottomrule\end{tabular}
\end{table*}